\documentclass[10pt,twocolumn,letterpaper]{article}

\usepackage{iccv}
\usepackage{times}
\usepackage{epsfig}
\usepackage{graphicx}
\usepackage{amsmath}
\usepackage{amssymb}

\usepackage{subcaption}
\usepackage{multirow}
\usepackage{color, colortbl}
\usepackage{tablefootnote}
\usepackage{pifont}

\DeclareMathOperator*{\argmin}{argmin}

\definecolor{First}{gray}{0.6}
\definecolor{Second}{gray}{0.75}
\definecolor{Third}{gray}{0.85}
\usepackage[pagebackref=true,breaklinks=true,letterpaper=true,colorlinks,bookmarks=false]{hyperref}

\iccvfinalcopy % *** Uncomment this line for the final submission

\def\iccvPaperID{465} % *** Enter the ICCV Paper ID here

\ificcvfinal\fi
\begin{document}

%%%%%%%%% TITLE
\title{FastFCN: Rethinking Dilated Convolution in the Backbone for Semantic Segmentation}

\author{Huikai Wu,~~~Junge Zhang,~~~Kaiqi Huang\\
Institute of Automation, Chinese Academy of Sciences\\
{\tt\small\{huikai.wu, jgzhang, kaiqi.huang\}@nlpr.ia.ac.cn}
\and
Kongming Liang,~~~Yizhou Yu\\
Deepwise AI Lab\\
{\tt\small liangkongming@deepwise.com, yizhouy@acm.org}
}

\maketitle
%\thispagestyle{empty}

%%%%%%%%% ABSTRACT
\begin{abstract}
Modern approaches for semantic segmentation usually employ dilated convolutions in the backbone to extract high-resolution feature maps, which brings heavy computation complexity and memory footprint.
To replace the time and memory consuming dilated convolutions, we propose a novel joint upsampling module named Joint Pyramid Upsampling (JPU) by formulating the task of extracting high-resolution feature maps into a joint upsampling problem.
With the proposed JPU, our method reduces the computation complexity by more than three times without performance loss.
Experiments show that JPU is superior to other upsampling modules, which can be plugged into many existing approaches to reduce computation complexity and improve performance.
By replacing dilated convolutions with the proposed JPU module, our method achieves the state-of-the-art performance in Pascal Context dataset (mIoU of 53.13\%) and ADE20K dataset (final score of 0.5584) while running 3 times faster.
Code is available in \url{https://github.com/wuhuikai/FastFCN}.
\end{abstract}

%%%%%%%%% BODY TEXT
\section{Introduction}
%---------------------------------------------------------------------------------
\begin{figure*} 
\begin{center}
	\begin{subfigure}[b]{0.33\linewidth}
		\includegraphics[width=\linewidth]{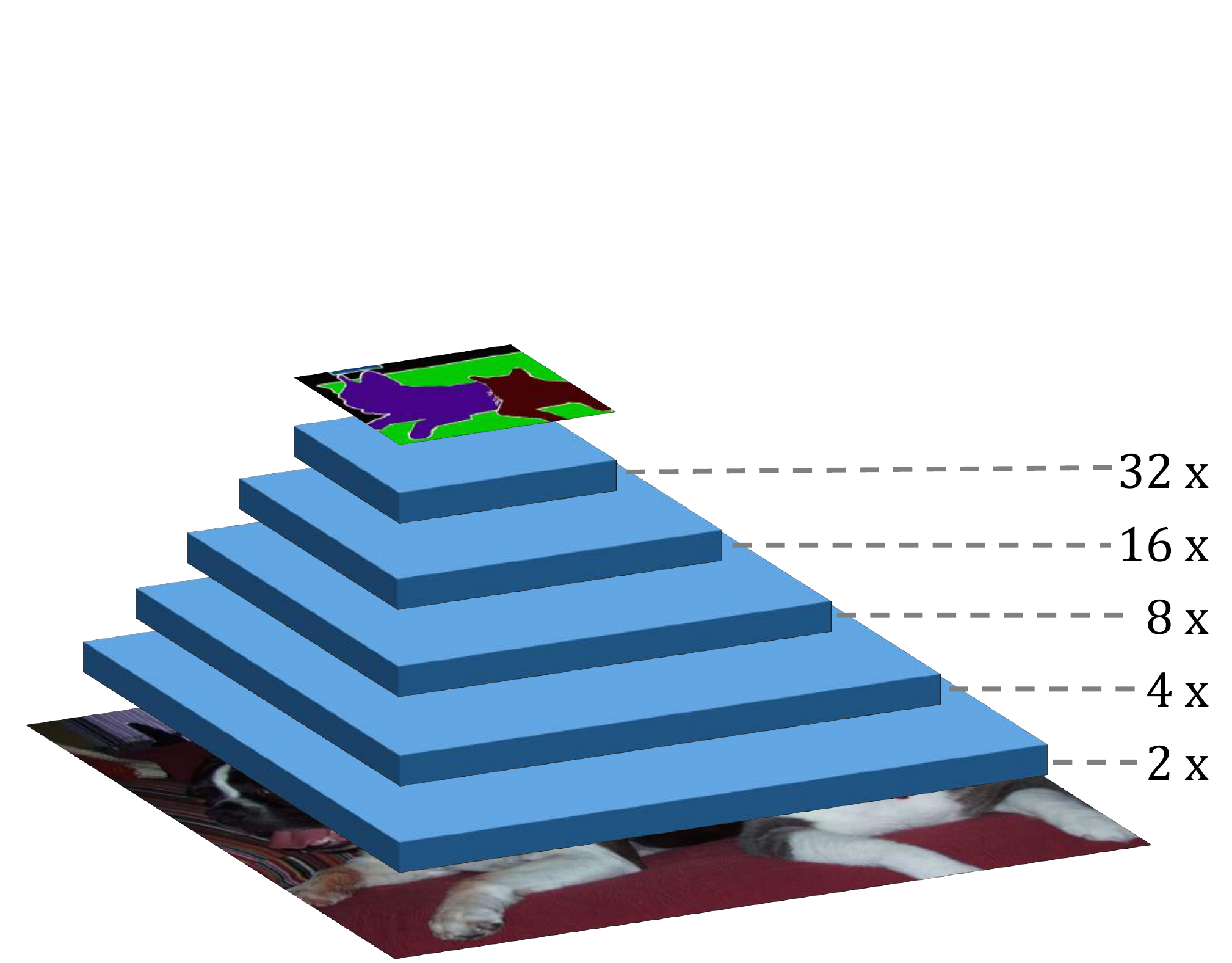}
      	\caption{FCN~\cite{long2015fully}}
      	\label{fig:fcns:fcn}
    \end{subfigure}
    \hfill
	\begin{subfigure}[b]{0.33\linewidth}
		\includegraphics[width=\linewidth]{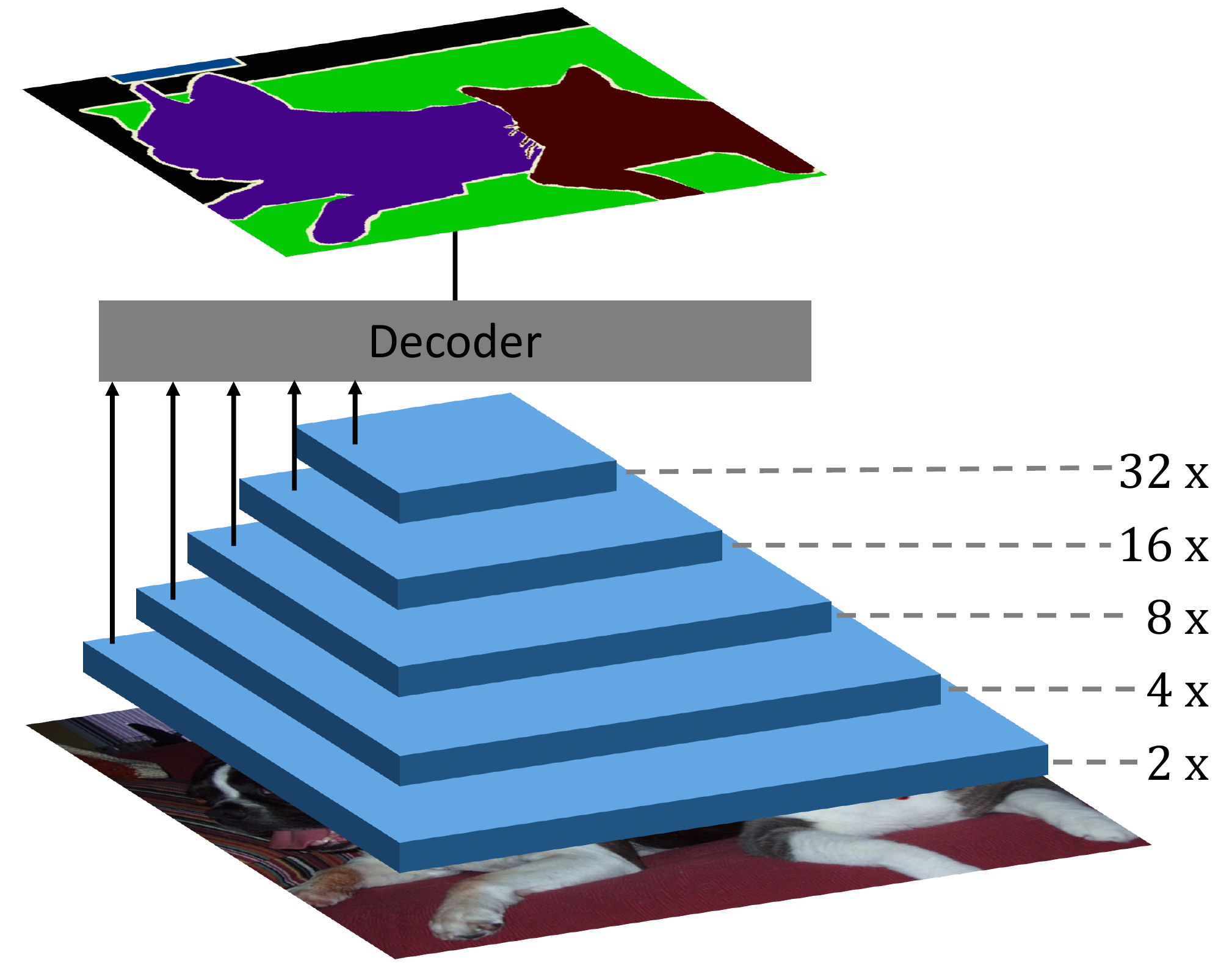}
      	\caption{EncoderDecoder}
      	\label{fig:fcns:encoder_decoder}
    \end{subfigure}
    \hfill
	\begin{subfigure}[b]{0.33\linewidth}
		\includegraphics[width=\linewidth]{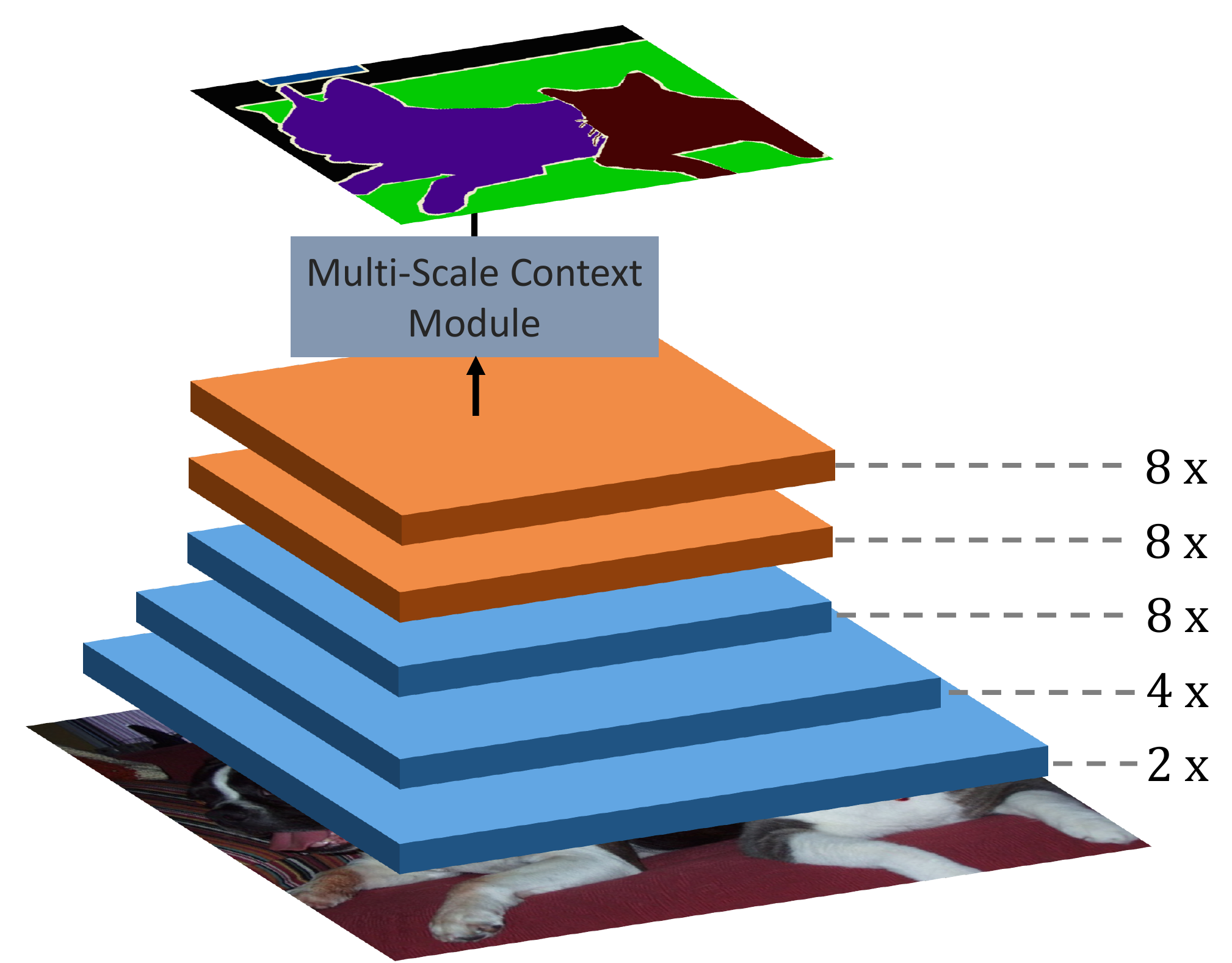}
      	\caption{DilatedFCN}
      	\label{fig:fcns:dilated_fcn}
    \end{subfigure}
\end{center}
	\caption{\textbf{Different types of networks for semantic segmentation.} (a) is the original FCN, (b) follows the encoder-decoder style, and (c) employs dilated convolutions to obtain high-resolution final feature maps. Best viewed in color.}
	\label{fig:fcns}
\end{figure*}
% ---------------------------------------------------------------------------------
Semantic segmentation~\cite{mottaghi2014role,zhou2017scene,caesar2018coco} is one of the fundamental tasks in computer vision, with the goal of assigning a semantic label to each pixel of an image.
Modern approaches usually employ a Fully Convolution Network (FCN)~\cite{long2015fully} to address this task, achieving tremendous success among several segmentation benchmarks.

The original FCN is proposed by Long \etal~\cite{long2015fully}, which is transformed from a Convolutional Neural Network (CNN)~\cite{lecun1998gradient,krizhevsky2012imagenet} designed for image classification.
Inheriting from the design for image classification, the original FCN downsamples the input image progressively by stride convolutions and/or spatial pooling layers, resulting in a final feature map in low resolution.
Although the final feature map encodes rich semantic information, the fine image structure information is lost, leading to inaccurate predictions around the object boundaries.
As shown in Figure~\ref{fig:fcns:fcn}, the original FCN typically downsamples the input image 5 times, reducing the spatial resolution of the final feature map by a factor of 32.

To obtain a high-resolution final feature map, \cite{badrinarayanan2015segnet,ronneberger2015u,lin2017refinenet,wojna2017devil,pohlen2017full} employ the original FCN as the encoder to capture high-level semantic information, and a decoder is designed to gradually recover the spatial information by combining multi-level feature maps from the encoder.
As shown in Figure~\ref{fig:fcns:encoder_decoder}, we term such methods EncoderDecoder, of which the final prediction generated by the decoder is in high resolution.
Alternatively, DeepLab~\cite{chen2018deeplab} removes the last two downsampling operations from the original FCN and introduces dilated (atrous) convolutions to maintain the receptive field of view unchanged.\footnote{In most cases, dilated convolutions in this paper refer to (1) removing downsampling operations and (2) replacing regular convolutions with dilated convolutions.}
Following DeepLab, \cite{zhao2017pyramid,chen2017rethinking,zhang2018context} employ a multi-scale context module on top of the final feature map, outperforming most EncoderDecoder methods significantly on several segmentation benchmarks.
As shown in Figure~\ref{fig:fcns:dilated_fcn}, the spatial resolution of the last feature map in DilatedFCN is 4 times larger than that in the original FCN, thus maintaining more structure and location information.

The dilated convolutions play an important role in maintaining the spatial resolution of the final feature map, leading to superior performance compared to most methods in EncoderDecoder.
However, the introduced dilated convolutions bring heavy computation complexity and memory footprint, which limit the usage in many real-time applications.
Taking ResNet-101~\cite{he2016deep} as an example, compared to the original FCN, 23 residual blocks (69 convolution layers) in DilatedFCN require to take 4 times more computation resources and memory usages, and 3 residual blocks (9 convolution layers) need to take 16 times more resources.

We aim at tackling the aforementioned issue caused by dilated convolutions in this paper.
To achieve this, we propose a novel joint upsampling module to replace the time and memory consuming dilated convolutions, namely Joint Pyramid Upsampling (JPU).
As a result, our method employs the original FCN as the backbone while applying JPU to upsample the low-resolution final feature map with output stride (OS) 32, resulting in a high-resolution feature map (OS=8).
Accordingly, the computation time and memory footprint of the whole segmentation framework is dramatically reduced.
Meanwhile, there's no performance loss when replacing the dilated convolutions with the proposed JPU.
We attribute this to the ability of JPU to exploit multi-scale context across multi-level feature maps.

To validate the effectiveness of our method, we first conduct a systematical experiment, showing that the proposed JPU can replace dilated convolutions in several popular approaches without performance loss.
We then test the proposed method on several segmentation benchmarks.
Results show that our method achieves the state-of-the-art performance while running more than 3 times faster.
Concretely, we outperform all the baselines on Pascal Context dataset~\cite{mottaghi2014role} by a large margin, which achieves the state-of-the-art performance with mIoU of 53.13\%.
On ADE20K dataset~\cite{zhou2017scene}, we obtain the mIoU of 42.75\% with ResNet-50 as the backbone, which sets a new record on the \textit{val} set.
Moreover, our method with ResNet-101 achieves the state-of-the-art performance in the \textit{test} set of ADE20K dataset.

In summary, our contributions are three folds, which are:
(1) We propose a computationally efficient joint upsampling module named JPU to replace the time and memory consuming dilated convolutions in the backbone.
(2) Based on the proposed JPU, the computation time and memory footprint of the whole segmentation framework can be reduced by a factor of more than 3 and meanwhile achieves better performance.
(3) Our method achieves the new state-of-the-art performance in both Pascal Context dataset (mIoU of 53.13\%) and ADE20K dataset (mIoU of 42.75\% with ResNet-50 as the backbone on the \textit{val} set and final score of 0.5584 with ResNet-101 on the \textit{test} set).
% ---------------------------------------------------------------------------------
\begin{figure*} 
\begin{center}
	\includegraphics[width=\linewidth]{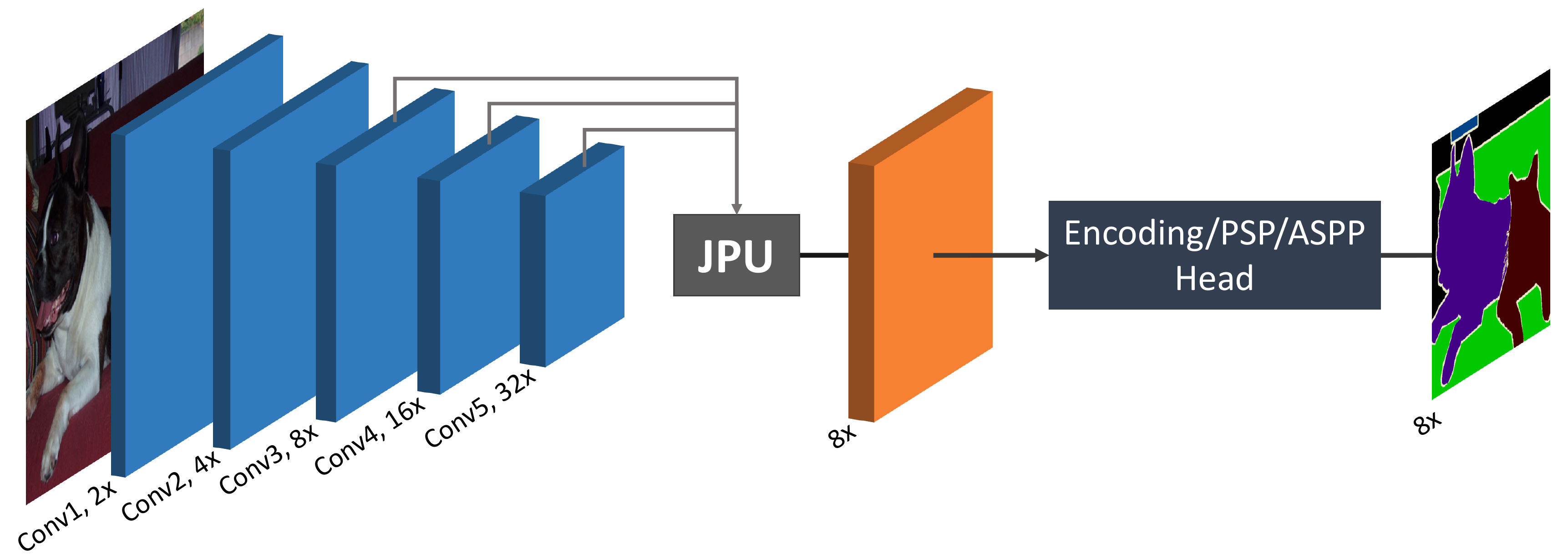}
\end{center}
	\caption{\textbf{Framework Overview of Our Method.} Our method employs the same backbone as the original FCN. After the backbone, a novel upsampling module named Joint Pyramid Upsampling (JPU) is proposed, which takes the last three feature maps as the inputs and generates a high-resolution feature map. A multi-scale/global context module is then employed to produce the final label map. Best viewed in color.}
	\label{fig:framework}
\end{figure*}
% ---------------------------------------------------------------------------------
\section{Related Work}
In this section, we first give an overview on methods for semantic segmentation, which can be categorized into two directions.
We then introduce some related works on upsampling.
% ---------------------------------------------------------------------------------
\subsection{Semantic Segmentation}
FCNs~\cite{long2015fully} have achieved huge success in semantic segmentation.
Following FCN, there're two prominent directions, namely DilatedFCN and EncoderDecoder.
DilatedFCNs~\cite{farabet2013learning,yu2015multi,chen2016attention,chen2017rethinking,zhao2017pyramid,zhang2018context,chen2018deeplab} utilize dilated convolutions to keep the receptive field of view and employ a multi-scale context module to process high-level feature maps.
Alternatively, EncoderDecoders~\cite{noh2015learning,ronneberger2015u,lin2017refinenet,amirul2017gated,peng2017large,fu2017stacked,yu2018learning,zhang2018exfuse} propose to utilize an encoder to extract multi-level feature maps, which are then combined into the final prediction by a decoder.
% ---------------------------------------------------------------------------------
\vspace{-1em}
\paragraph{DilatedFCN}
In order to capture multi-scale context information on the high-resolution final feature map, PSPNet~\cite{zhao2017pyramid} performs pooling operations at multiple grid scales while DeepLabV3~\cite{chen2017rethinking} employs parallel atrous convolutions with different rates named ASPP.
Alternatively, EncNet~\cite{zhang2018context} utilizes the Context Encoding Module to capture global contextual information.
Differently, our method proposes a joint upsampling module named JPU to replace the dilated convolutions in the backbone of DilatedFCNs, which can reduce computation complexity dramatically without performance loss.
% ---------------------------------------------------------------------------------
\vspace{-1em}
\paragraph{EncoderDecoder}
To gradually recover the spatial information, \cite{ronneberger2015u} introduces skip connections to construct U-Net, which combines the encoder features and the corresponding decoder activations.
\cite{lin2017refinenet} proposes a multi-path refinement network, which explicitly exploits all the information available along the down-sampling process.
DeepLabV3+~\cite{chen2018encoder} combines the advantages of DilatedFCN and EncoderDecoder, which employs DeepLabV3 as the encoder.
Our method is complementary to DeepLabV3+, which can reduce the computation overload of DeepLabV3 without performance loss.
% ---------------------------------------------------------------------------------
\subsection{Upsampling}
In our method, we propose a module to upsample a low-resolution feature map given high-resolution feature maps as guidance, which is closely related to joint upsampling as well as data-dependent upsampling.
\vspace{-1em}
\paragraph{Joint Upsampling}
In the literature of image processing, joint upsampling aims at leveraging the guidance image as a prior and transferring the structural details from the guidance image to the target image.
\cite{li2016deep} constructs a joint filter based on CNNs, which learns to recover the structure details in the guidance image.
\cite{wu2018fast} proposes an end-to-end trainable guided filtering module, which upsamples a low-resolution image conditionally.
Our method is related to the aforementioned approaches.
However, the proposed JPU is designed for processing feature maps with a large number of channels while \cite{li2016deep,wu2018fast} are specially designed for processing 3-channel images, which fail to capture the complex relations in high dimensional feature maps.
Besides, the motivation and target of our method is completely different.
\vspace{-1em}
\paragraph{Data-Dependent Upsampling}
DUpsampling~\cite{tian2019decoders} is also related to our method, which takes advantages of the redundancy in the segmentation label space and is able to recover the pixel-wise prediction from low-resolution outputs of CNNs.
Compared to our method, DUpsampling has a strong dependency on the label space, which generalizes poorly to a larger or more complex label space.
% ---------------------------------------------------------------------------------
\section{Method}
% ---------------------------------------------------------------------------------
In this section, we first introduce the most popular methods for semantic segmentation, named DilatedFCNs.
We then reform the architecture of DilatedFCNs with a novel joint upsampling module, Joint Pyramid Upsampling (JPU).
Finally, we discuss the proposed JPU in details, before which joint upsampling, dilated convolution and stride convolution are briefly introduced.
% ---------------------------------------------------------------------------------
\subsection{DilatedFCN}
To exploit Deep CNNs in semantic segmentation, Long~\etal~\cite{long2015fully} transform the CNN designed for image classification into FCN.
Taking ResNet-101 as an example, the original CNN contains 5 convolution stages, a global average pooling layer and a linear layer.
To construct an FCN, the global average pooling layer and the linear layer are replaced by a convolution layer, which is used to generate the final label map, as shown in Figure~\ref{fig:fcns:fcn}.
Between each two consecutive convolution stages, stride convolutions and/or spatial pooling layers are employed, resulting in 5 feature maps with gradually reduced spatial resolutions.

The spatial resolution of the last feature map in FCN is reduced by a factor of 32, leading to inaccurate predictions about the locations and details.
To obtain a final feature map with high resolution, DeepLab~\cite{chen2018deeplab} removes the downsampling operations before the last two feature maps, as shown in Figure~\ref{fig:fcns:dilated_fcn}.
Besides, the convolution layers inside the last two convolution stages are replaced by dilated convolutions to maintain the receptive field of view, thus named DilatedFCN.
As a result, the resolution of the last feature map is reduced by a factor of 8, which reserves more location and detail information.
Following DeepLab, \cite{zhao2017pyramid,chen2017rethinking} propose a multi-scale context module to capture context information from the last feature map, achieving tremendous success in several segmentation benchmarks.
% ---------------------------------------------------------------------------------
\begin{figure*} 
\begin{center}
	\begin{subfigure}[b]{0.608\linewidth}
		\includegraphics[width=\linewidth]{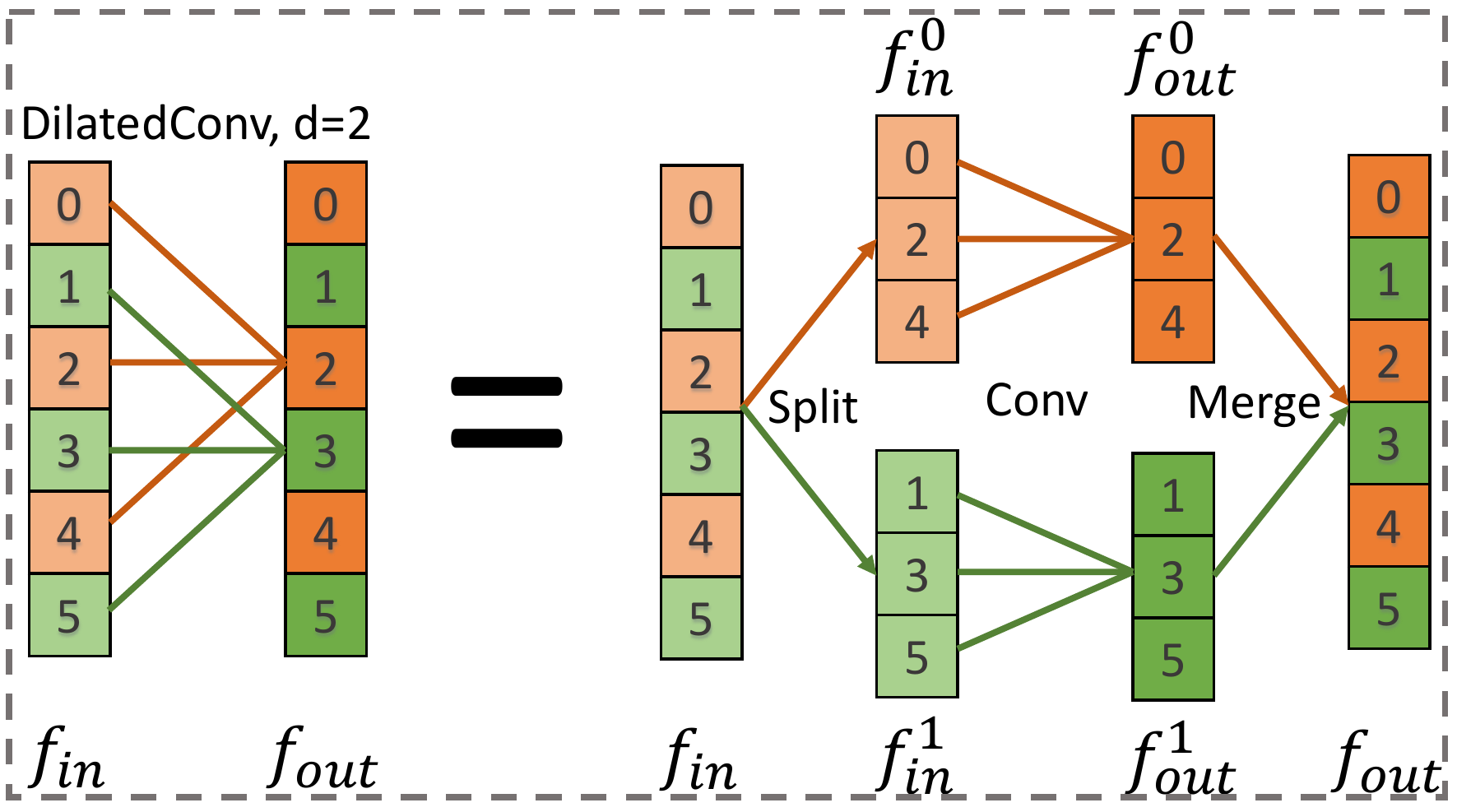}
      	\caption{Dilated Convolution in 1D (d=2)}
      	\label{fig:dilation:dilated_conv}
    \end{subfigure}
    \hfill
    \begin{subfigure}[b]{0.372\linewidth}
		\includegraphics[width=\linewidth]{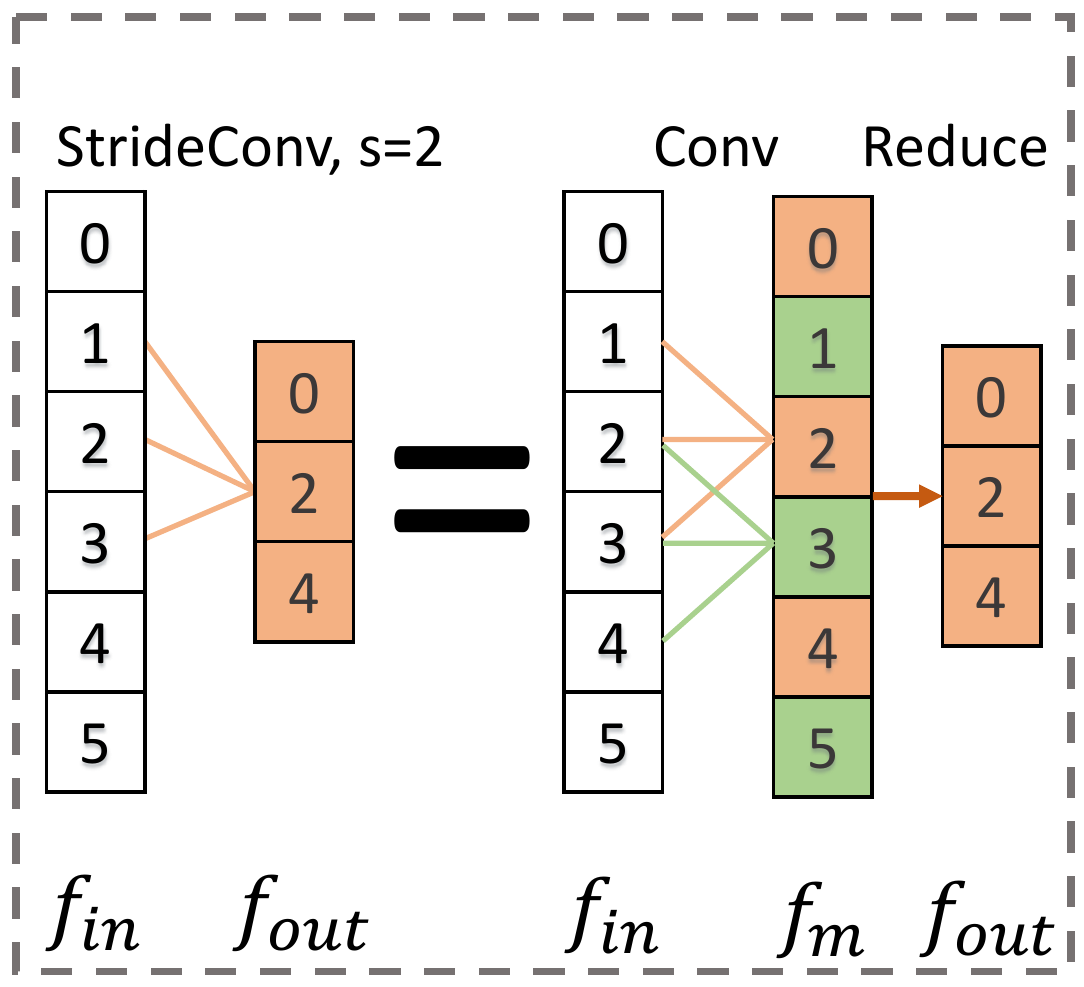}
      	\caption{Stride Convolution in 1D (s=2)}
      	\label{fig:dilation:stride_conv}
    \end{subfigure}
\end{center}
	\caption{Dilated Convolution (dilation rate=2) and Stride Convolution (stride=2) in 1D. Best viewed in color.}
	\label{fig:dilation}
\end{figure*}
% ---------------------------------------------------------------------------------
\subsection{The Framework of Our Method}
To obtain a high-resolution final feature map, methods in DilatedFCN remove the last two downsampling operations from the original FCN, which bring in heavy computation complexity and memory footprint due to the enlarged feature maps.
In this paper, we aim at seeking an alternative way to approximate the final feature map of DilatedFCN without computation and memory overload.
Meanwhile, we expect the performance of our method to be as good as that of the original DilatedFCNs.

To achieve this, we first put back all the stride convolutions removed by DilatedFCN, while replacing all the dilated convolutions with regular convolution layers.
As shown in Figure~\ref{fig:framework}, the backbone of our method is the same as that of the original FCN, where the spatial resolutions of the five feature maps (Conv1$-$Conv5) are gradually reduced by a factor of 2.
To obtain a feature map similar to the final feature map of DilatedFCN, we propose a novel module named Joint Pyramid Upsampling (JPU), which takes the last three feature maps (Conv3$-$Conv5) as inputs.
Then a multi-scale context module (PSP~\cite{zhao2017pyramid}/ASPP~\cite{chen2017rethinking}) or a global context module (Encoding~\cite{zhang2018context}) is employed to produce the final predictions.
 
Compared to DilatedFCN, our method takes 4 times fewer computation and memory resources in 23 residual blocks (69 layers) and 16 times fewer in 3 blocks (9 layers) when the backbone is ResNet-101.
Thus, our method runs much faster than DilatedFCN while consuming less memory.
% ---------------------------------------------------------------------------------
\subsection{Joint Pyramid Upsampling}
The proposed JPU is designed for generating a feature map that approximates the activations of the final feature map from the backbone of DilatedFCN.
Such a problem can be reformulated into joint upsampling, which is then resolved by a CNN designed for this task.
\subsubsection{Background}
\paragraph{Joint Upsampling}
Given a low-resolution target image and a high-resolution guidance image, joint upsampling aims at generating a high-resolution target image by transferring details and structures from the guidance image.
Generally, the low-resolution target image $y_l$ is generated by employing a transformation $f(\cdot)$ on the low-resolution guidance image $x_l$, \ie $y_l = f(x_l)$.
Given $x_l$ and $y_l$, we are required to obtain a transformation $\hat{f}(\cdot)$ to approximate $f(\cdot)$, where the computation complexity of $\hat{f}(\cdot)$ is much lower than $f(\cdot)$.
For example, if $f(\cdot)$ is a multi-layer perceptron (MLP), then $\hat{f}(\cdot)$ can be simplified as a linear transformation.
The high-resolution target image $y_h$ is then obtained by applying $\hat{f}(\cdot)$ on the high-resolution guidance image $x_h$, \ie $y_h = \hat{f}(x_h)$.
Formally, given $x_l$, $y_l$ and $x_h$, joint upsampling is defined as follows:
\begin{equation}
	y_h = \hat{f}(x_h),~\text{where} ~\hat{f}(\cdot) = \argmin_{h(\cdot)\in \mathcal{H}} ||y_l - h(x_l)||,
	\label{eq:ju}
\end{equation}
where $\mathcal{H}$ is a set of all possible transformation functions, and $||\cdot||$ is a pre-defined distance metric.
\vspace{-1em}
\paragraph{Dilated Convolution}
Dilated convolution is introduced in DeepLab~\cite{chen2018deeplab} for obtaining high-resolution feature maps while maintaining the receptive field of view.
Figure~\ref{fig:dilation:dilated_conv} gives an illustration of the dilated convolution in 1D (dilation rate = 2), which can be divided into the following three steps:
(1) split the input feature $f_{in}$ into two groups $f_{in}^0$ and $f_{in}^1$ according to the parity of the index,
(2) process each feature with the same convolution layer, resulting in $f_{out}^0$ and $f_{out}^1$,
and (3) merge the two generated features interlaced to obtain the output feature $f_{out}$.
\vspace{-1em}
\paragraph{Stride Convolution}
Stride convolution is proposed to transform the input feature into an output feature with reduced spatial resolution, which is equivalent to the following two steps as shown in Figure~\ref{fig:dilation:stride_conv}:
(1) process the input feature $f_{in}$ with a regular convolution to obtain the intermediate feature $f_m$,
and (2) remove the elements with an odd index, resulting in $f_{out}$.

\subsubsection{Reformulating into Joint Upsampling}
\quad The differences between the backbone of our method and DilatedFCN lie on the last two convolution stages.
Taking the 4th convolution stage (Conv4) as an example, in DilatedFCN, the input feature map is first processed by a regular convolution layer, followed by a series of dilated convolutions (d=2).
Differently, our method first processes the input feature map with a stride convolution (s=2), and then employs several regular convolutions to generate the output.

Formally, given the input feature map $x$, the output feature map $y_{d}$ in DilatedFCN is obtained as follows:
\begin{equation}
	\begin{aligned}
		y_d &= x\to C_r \to \underbrace{C_d\to ...... \to C_d}_{n} \\
		    &= x\to C_r \to \underbrace{SC_rM\to ...... \to SC_rM}_{n}~\text{(Fig~\ref{fig:dilation:dilated_conv})}\\
		    &= x\to C_r \to S \to \underbrace{C_r\to ...... \to C_r}_{n}\to M \\
		    &= y_m \to S \to C_r^n \to M\\
		    &= \{y_m^{0}, y_m^{1}\}\to C_r^n \to M~\text{(Fig~\ref{fig:dilation:dilated_conv})},
	\end{aligned}
	\label{eq:dilated:conv}
\end{equation}
while in our method, the output feature map $y_{s}$ is generated as follows:
\begin{equation}
	\begin{aligned}
		y_s &= x\to C_s \to \underbrace{C_r\to ...... \to C_r}_{n}\\
		    &= x\to C_r \to R \to \underbrace{C_r\to ...... \to C_r}_{n}~\text{(Fig~\ref{fig:dilation:stride_conv})}\\
		    &= y_m \to R \to C_r^n = y_m^{0}\to C_r^n~\text{(Fig~\ref{fig:dilation:stride_conv})}.
	\end{aligned}
	\label{eq:ours:conv}
\end{equation}
$C_r$, $C_d$, and $C_s$ represent a regular/dilated/stride convolution respectively, and $C_r^n$ is $n$ layers of regular convolutions.
$S$, $M$ and $R$ are split, merge, and reduce operations in Figure~\ref{fig:dilation}, where adjacent $S$ and $M$ operations can be canceled out.
Notably, the convolutions in Equations~\ref{eq:dilated:conv} and \ref{eq:ours:conv} are in 1D, which is for simplicity.
Similar results can be obtained for 2D convolutions.

The aforementioned equations show that $y_s$ and $y_d$ can be obtained with the same function $C_r^n$ with different inputs: $y_m^0$ and $y_m$, where the former is downsampled from the latter.
Thus, given $x$ and $y_s$, the feature map $y$ that approximates $y_d$ can be obtained as follows:
\begin{equation}
	\begin{aligned}
		y = &\{y_m^{0}, y_m^{1}\}\to \hat{h}\to M \\
		    &\text{where} ~\hat{h} = \argmin_{h\in \mathcal{H}} ||y_s - h(y_m^{0})||,\\
		    &\quad\quad~~~y_m = x\to C_r,
	\end{aligned}
	\label{eq:jpu}
\end{equation}
which is the same as the joint upsampling problem defined in Equation~\ref{eq:ju}.
Similar conclusions can be easily obtained for the 5th convolution stage (Conv5).
% ---------------------------------------------------------------------------------
\subsubsection{Solving with CNNs}
\quad Equation~\ref{eq:jpu} is an optimization problem, which takes lots of time to converge through the iterative gradient descent.
Alternatively, we propose to approximate the optimization process with a CNN module.
To achieve this, we first require to generate $y_m$ given $x$, as shown in Equation~\ref{eq:jpu}.
Then, features from $y_m^0$ and $y_s$ need to be gathered for learning the mapping $\hat{h}$.
Finally, a convolution block is required to transform the gathered features into the final prediction $y$.

\begin{figure} 
\begin{center}
	\includegraphics[width=\linewidth]{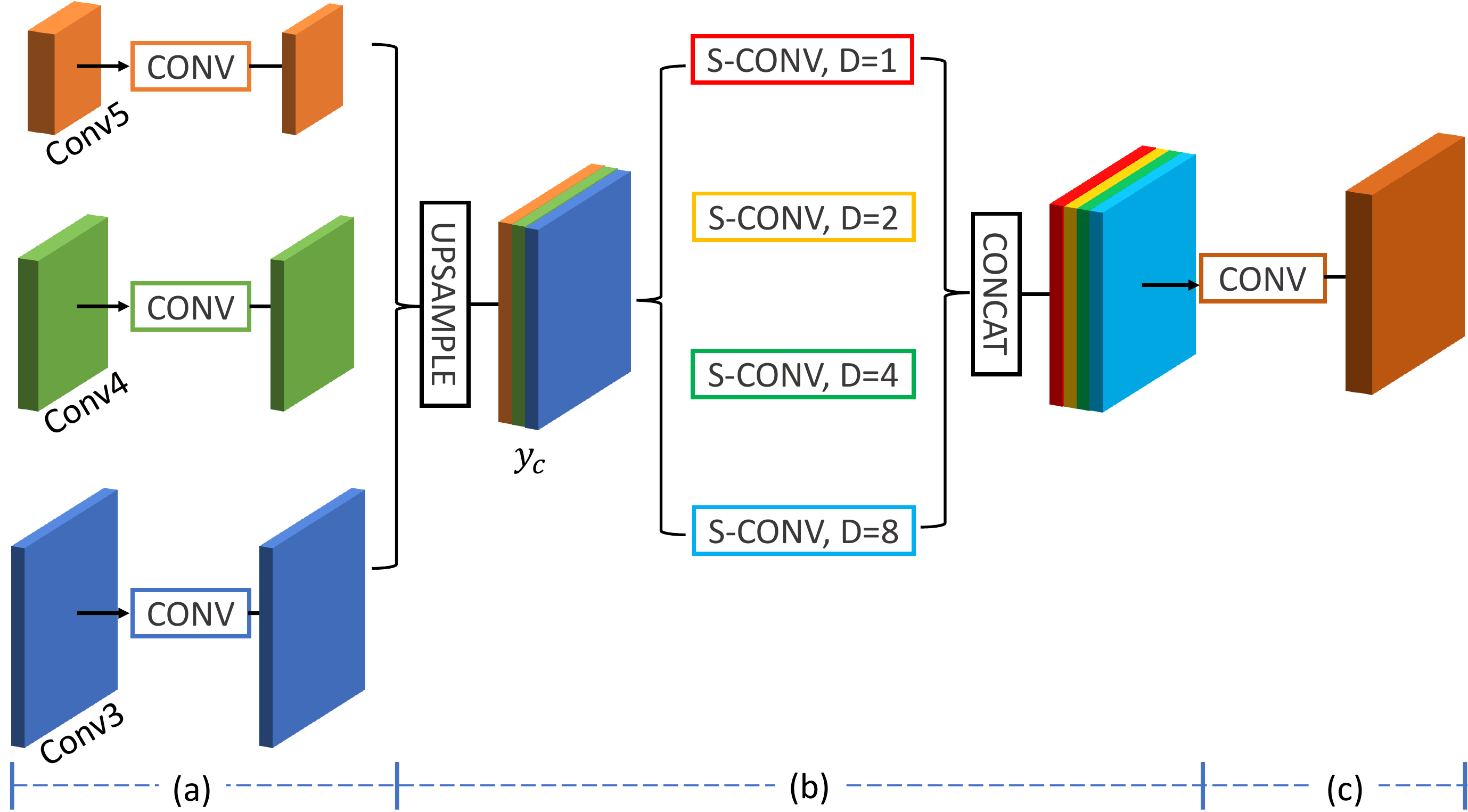}
\end{center}
	\caption{\textbf{The Proposed Joint Pyramid Upsampling (JPU).} Best viewed in color.}
	\label{fig:jpu}
\end{figure}

Following the aforementioned analysis, we design the JPU module as in Figure~\ref{fig:jpu}. 
Concretely, each input feature map is firstly processed by a regular convolution block (Fig.~\ref{fig:jpu}a), which is designed for (1) generating $y_m$ given $x$, and (2) transforming $f_m$ into an embedding space with reduced dimensions.
As a result, all the input features are mapped into the same space, which enables a better fusion and reduces the computation complexity.

\begin{figure} 
\begin{center}
	\includegraphics[width=\linewidth]{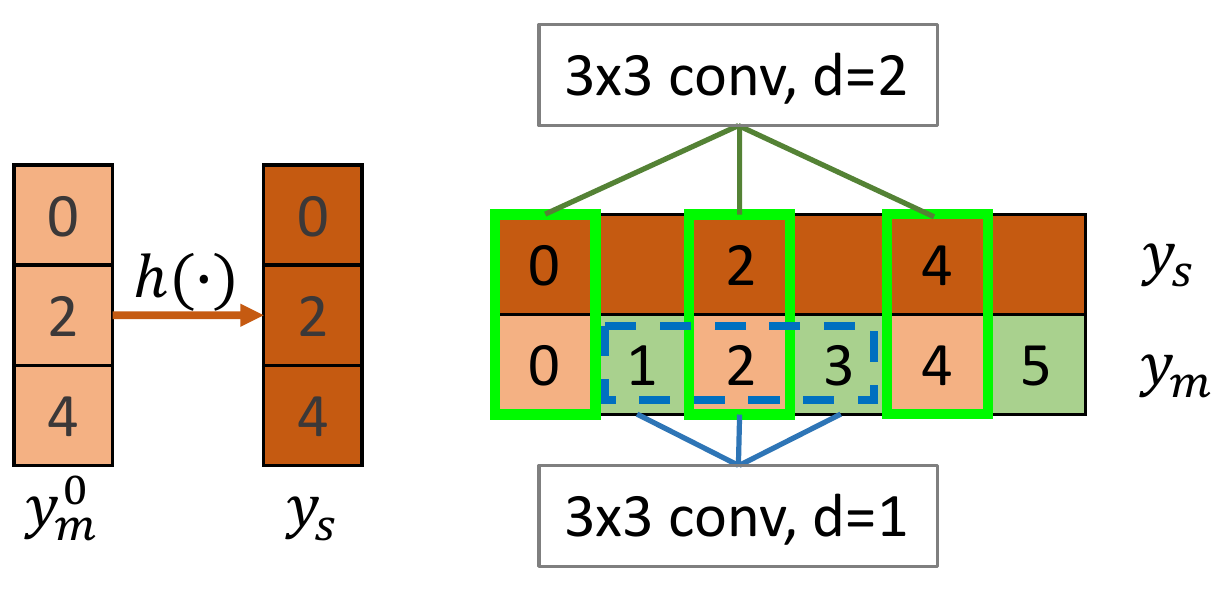}
\end{center}
	\caption{The convolution with dilation rate 1 focuses on $y_m^0$ and the rest part of $y_m$, and the convolution with dilation rate 2 aims at $y_m^0$ and $y_s$. Best viewed in color.}
	\label{fig:aspp}
\end{figure}

Then, the generated feature maps are upsampled and concatenated, resulting in $y_c$ (Fig.~\ref{fig:jpu}b).
Four separable convolutions~\cite{howard2017mobilenets,chollet2017xception} with different dilation rates (1, 2, 4, and 8) are employed in parallel to extract features from $y_c$, where different dilation rates take different functions.
Concretely, the convolution with dilation rate 1 is employed to capture the relation between $y_m^{0}$ and the rest part of $y_m$, as shown by the blue box in Figure~\ref{fig:aspp}.
Alternatively, the convolutions with dilation rate 2, 4 and 8 are designed for learning the mapping $\hat{h}$ to transform $y_m^{0}$ into $y_s$, as shown by the green boxes in Figure~\ref{fig:aspp}.
Thus, JPU can extract multi-scale context information from multi-level feature maps, which leads to a better performance.
This is significantly different from ASPP~\cite{chen2017rethinking}, which only exploit the information in the last feature map.

The extracted features encode the mapping between $y_m^{0}$ and $y_s$ as well as the relation between $y_m^{0}$ and the rest part of $y_m$.
Thus, another regular convolution block is employed, which transforms the features into the final predictions (Fig.~\ref{fig:jpu}c).

Notably, the proposed JPU module solves two closely related joint upsampling problems jointly, which are (1) upsampling Conv4 based on Conv3 (the 4th convolution stage), and (2) upscaling Conv5 with the guidance of the enlarged Conv4 (the 5th convolution stage).
% ---------------------------------------------------------------------------------
% ---------------------------------------------------------------------------------
% ---------------------------------------------------------------------------------
\section{Experiment}
In this section, we first introduce the datasets used in our experiments as well as the implementation details.
We then conduct a systematic ablation study to show the effectiveness of the proposed JPU from the view of both performance and efficiency.
Finally, to compare with the state-of-the-art methods, we report the performance on two segmentation datasets, Pascal Context~\cite{mottaghi2014role} and ADE20K~\cite{zhou2017scene}, which are widely used as the segmentation benchmarks.
Moreover, we also show some visual results to demonstrate the superiority of our method.
% ---------------------------------------------------------------------------------
\subsection{Experimental Settings}
% ---------------------------------------------------------------------------------
\paragraph{Dataset}
Pascal Context dataset~\cite{mottaghi2014role} is based on the PASCAL VOC 2010 detection challenge, which provides additional pixel-wise semantic annotations.
There're 4,998 images for training (\textit{train}) and 5,105 images for testing (\textit{val}).
Following the prior works~\cite{lin2017refinenet,chen2018deeplab,zhang2018context}, we use the most frequent 59 object categories plus background (60 classes in total) as the semantic labels.
% ---------------------------------------------------------------------------------
\vspace{-1em}
\paragraph{Implementation Details}
Our method is implemented in PyTorch~\cite{paszke2017automatic}.
For training on Pascal Context, we follow the protocol presented in~\cite{zhang2018context}.
Concretely, we set the learning rate to 0.001 initially, which gradually decreases to 0 by following the "poly" strategy (power = 0.9).
For data augmentation, we randomly scale (from 0.5 to 2.0) and left-right flip the input images.
The images are then cropped to $480\times 480$ and grouped with batch size 16.
The network is trained for 80 epochs with SGD, of which the momentum is set to 0.9 and weight decay is set to 1e-4.
All the experiments are conducted in a workstation with 4 Titan-Xp GPUs (12G per GPU).
We employ pixel-wise cross-entropy as the loss function.
ResNet-50 and ResNet-101 are used as the backbone, which are widely used in most existing segmentation methods as the standard backbones.
% ---------------------------------------------------------------------------------
\subsection{Ablation Study}
To show the effectiveness of the proposed method, we conduct a systematical ablation study on Pascal Context dataset with ResNet-50 as the backbone, as shown in Table~\ref{table:ablation}.
We report the standard evaluation metrics of pixel accuracy (pixAcc) and mean Intersection of Union (mIoU).
Notably, no multi-scale testing and left-right flipping are applied to the \textit{val} images.
% ---------------------------------------------------------------------------------
\vspace{-1em}
\paragraph{Dilated Convolutions}
For methods in DilatedFCN, the downsampling operations in the last two convolution stages are removed, resulting in the output stride (OS) to be 8.
Encoding-8-None in Table~\ref{table:ablation} represents the original EncNet~\cite{zhang2018context}.
To show the effect of dilated convolutions, we replace the backbone of EncNet with that of the original FCN (the same as our method), resulting in the OS to be 32.
We then upsample the last feature map by 4 times with bilinear interpolation before feeding it into the Encoding Head, noted as Encoding-32-Bilinear.
As shown in Table~\ref{table:ablation}, Encoding-32-Bilinear performs significantly worse than Encoding-8-None, which shows that it's not trivial to replace the dilated convolutions in the backbone of DilatedFCNs.
% ---------------------------------------------------------------------------------
\vspace{-1em}
\paragraph{Upsampling Module}
To show the effectiveness of the proposed JPU, we compare it with other classic upsampling methods, bilinear upsampling and feature pyramid network (FPN)~\cite{lin2017feature}.
As shown in Table~\ref{table:ablation}, FPN outperforms bilinear interpolation by a large margin.
Even compared with EncNet, FPN achieves comparable performance in both pixAcc and mIoU.
By replacing FPN with our JPU, our method outperforms both FPN and EncNet by more than 1\% in mIoU, which achieves the state-of-the-art performance.

\begin{figure} 
\begin{center}
	\begin{subfigure}[b]{0.32\linewidth}
		\includegraphics[width=\linewidth]{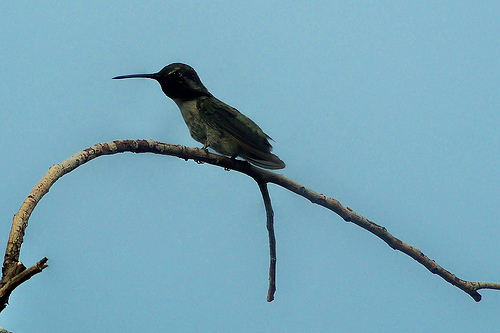}
      	\caption{Input}
      	\label{fig:pcontext_res50:input}
    \end{subfigure}
    \hfill
	\begin{subfigure}[b]{0.32\linewidth}
		\includegraphics[width=\linewidth]{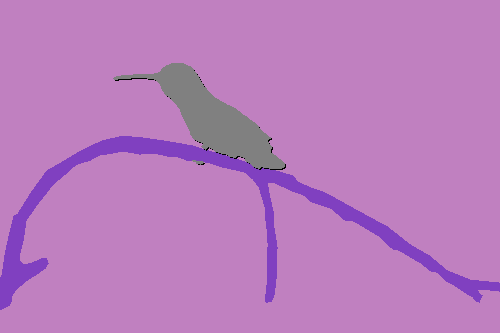}
      	\caption{GT}
      	\label{fig:pcontext_res50:gt}
    \end{subfigure}
	\hfill
	\begin{subfigure}[b]{0.32\linewidth}
		\includegraphics[width=\linewidth]{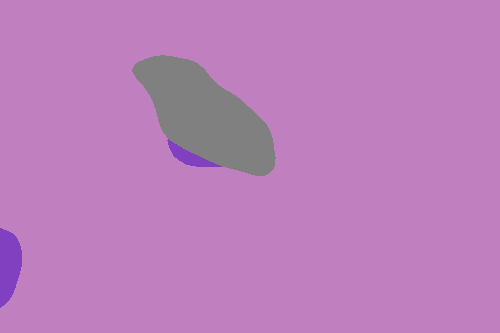}
      	\caption{Bilinear}
      	\label{fig:pcontext_res50:bilinear}
    \end{subfigure}\\
    \begin{subfigure}[b]{0.32\linewidth}
		\includegraphics[width=\linewidth]{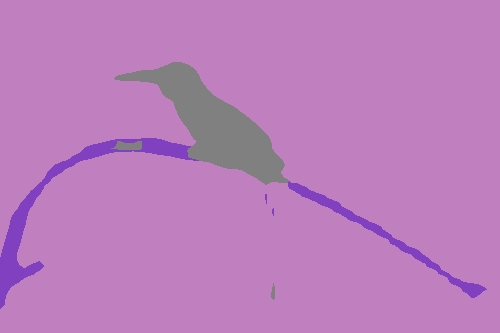}
      	\caption{FPN~\cite{lin2017feature}}
      	\label{fig:pcontext_res50:fpn}
    \end{subfigure}
    \hfill
	\begin{subfigure}[b]{0.32\linewidth}
		\includegraphics[width=\linewidth]{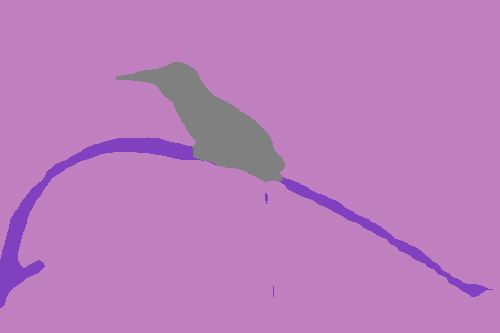}
      	\caption{EncNet~\cite{zhang2018context}}
      	\label{fig:pcontext_res50:encnet}
    \end{subfigure}
	\hfill
	\begin{subfigure}[b]{0.32\linewidth}
		\includegraphics[width=\linewidth]{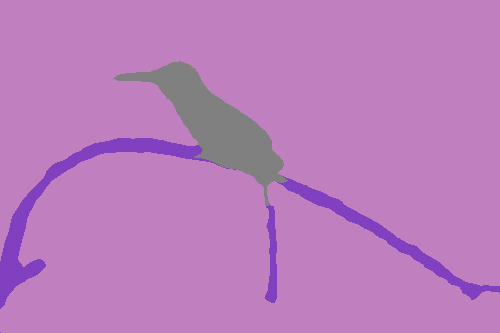}
      	\caption{Ours}
      	\label{fig:pcontext_res50:ours}
    \end{subfigure}
\end{center}
	\caption{Visual comparison of different upsampling modules with Encoding Head and ResNet-50 as the backbone.}
	\label{fig:pcontext_res50}
\end{figure}

The visual results are shown in Figure~\ref{fig:pcontext_res50}.
Encoding-32-Bilinear (Fig.~\ref{fig:pcontext_res50:bilinear}) captures the global semantic information successfully, which gives a rough segmentation of the bird and sky.
However, the boundary of the bird is inaccurate, and most parts of the branch are failed to be labeled out.
When replacing bilinear interpolation with FPN (Fig.~\ref{fig:pcontext_res50:fpn}), the bird and branch are labeled out successfully with accurate boundaries, which shows the effect of combing low-level and high-level feature maps.
A slightly better result can be obtained with dilated convolutions (Fig.~\ref{fig:pcontext_res50:encnet}).
As for our method (Fig.~\ref{fig:pcontext_res50:ours}), it labels out both the main branch and the side shoot accurately, which shows the effectiveness of the proposed joint upsampling module.
Particularly, the side shoot demonstrates the ability of JPU to extract multi-scale context from multi-level feature maps.
Thus, our method can achieve a better performance.
% ---------------------------------------------------------------------------------
\vspace{-1em}
\paragraph{Generalization to Other Methods}
To show the generalization ability of the proposed JPU, we replace EncNet with two popular methods in DilatedFCN, namely DeepLabV3 (ASPP Head)~\cite{chen2017rethinking} and PSPNet~\cite{zhao2017pyramid}.
As shown in Table~\ref{table:ablation}, our methods transformed from DeepLabV3 and PSP outperforms the corresponding original methods consistently.
% ---------------------------------------------------------------------------------
\begin{table}
\begin{center}
\begin{tabular}{l|c|l|cc}
\hline
Head & OS & Upsampling & pixAcc\% & mIoU\% \\
\hline
\hline
\multirow{4}{*}{Encoding~\cite{zhang2018context}} & 8 & None & 78.39 & 49.91\\
\cline{2-2}
 & \multirow{3}{*}{32} & Bilinear & 76.10 & 46.47\\
 &  & FPN~\cite{lin2017feature} & 78.16 & 49.59\\
 &  & JPU (ours) & \cellcolor{First}\textbf{78.98} & \cellcolor{First}\textbf{51.05}\\
\hline
\multirow{2}{*}{ASPP~\cite{chen2017rethinking}} & 8 & None & 78.27 & 49.19\\
 & 32 & JPU (ours) & \cellcolor{Third}78.79 & 50.07\\
\hline
\multirow{2}{*}{PSP~\cite{zhao2017pyramid}} & 8 & None & 78.60 & \cellcolor{Third}50.58\\
 & 32 & JPU (ours) & \cellcolor{Second}78.91 & \cellcolor{Second}50.89\\
\hline
\end{tabular}
\end{center}
	\caption{Performance on the \textit{val} set of Pascal Context dataset with the ResNet-50 as the backbone.}
	\label{table:ablation}
\end{table}
% ---------------------------------------------------------------------------------
% ---------------------------------------------------------------------------------
\begin{table}
\begin{center}
\begin{tabular}{l|l|l|c}
\hline
Backbone & Head & Upsampling & FPS\\
\hline
\multirow{8}{*}{ResNet-50} & \multirow{4}{*}{Encoding~\cite{zhang2018context}} & None & 18.77\\
 &  & Bilinear & \cellcolor{First}45.67\\
 &  & FPN~\cite{lin2017feature} & \cellcolor{Second}37.87\\
 &  & JPU (ours) & \cellcolor{Third}37.56\\
\cline{2-4}
 & \multirow{2}{*}{ASPP~\cite{chen2017rethinking}} & None & 15.99\\
 & & JPU (ours) & 20.67\\
\cline{2-4}
 & \multirow{2}{*}{PSP~\cite{zhao2017pyramid}} & None & 18.08\\
 &  & JPU (ours) & 28.48\\
\hline
\hline
\multirow{8}{*}{ResNet-101} & \multirow{4}{*}{Encoding~\cite{zhang2018context}} & None & 10.51\\
 & & Bilinear & \cellcolor{First}35.20\\
 & & FPN~\cite{lin2017feature} & \cellcolor{Second}32.40\\
 & & JPU (ours) & \cellcolor{Third}32.02\\
\cline{2-4}
 & \multirow{2}{*}{ASPP~\cite{chen2017rethinking}} & None & 10.46\\
 &  & JPU (ours) & 18.08\\
\cline{2-4}
 & \multirow{2}{*}{PSP~\cite{zhao2017pyramid}} & None & 11.36\\
 &  & JPU (ours) & 23.87\\ 
\hline
\end{tabular}
\end{center}
	\caption{\textbf{Comparison of Computation Complexity.} The FPS is measured on a Titan-Xp GPU with a $512\times 512$ image as input, which is averaged among 100 runs.}
	\label{table:speed}
\end{table}
% ---------------------------------------------------------------------------------
\vspace{-1em}
\paragraph{FPS}
To compare the computation complexity, we employ frame per second (FPS) as the evaluation metric, which is measured on a Titan-Xp GPU with a $512\times 512$ image as input.
As shown in Table~\ref{table:speed}, the reported FPS is averaged among 100 runs.
For ResNet-50, our method (Encoding-JPU) runs about two times faster than EncNet (Encoding-None).
When changing the backbone to ResNet-101, our method runs more than three times faster than EncNet.
The speed of our method is also comparable to FPN, but our method achieves much better performance.
As for DeepLabV3 (ASPP) and PSP, our method can accelerate them to a certain degree while having a better performance.
% ---------------------------------------------------------------------------------
\subsection{Comparison with Other Methods}
% ---------------------------------------------------------------------------------
\begin{table}
\begin{center}
\begin{tabular}{r|l|c}
\hline
Method & Backbone & mIoU\%\\
\hline
FCN-8s~\cite{long2015fully} & & 37.8\\
CRF-RNN~\cite{zheng2015conditional} & & 39.3\\
ParseNet~\cite{liu2015parsenet} & & 40.4\\
BoxSup~\cite{dai2015boxsup} & & 40.5\\
HO\_CRF~\cite{arnab2016higher} & & 41.3\\
Piecewise~\cite{lin2016efficient} & & 43.3\\
VeryDeep~\cite{wu2016bridging} & & 44.5\\
DeepLabV2~\cite{chen2018deeplab} & ResNet-101 + COCO & 45.7\\
RefineNet~\cite{lin2017refinenet} & ResNet-152 & 47.3\\
EncNet~\cite{zhang2018context} & ResNet-101 & \cellcolor{Third}51.7\\
DUpsampling~\cite{tian2019decoders} & Xception-71 & \cellcolor{Second}52.5\\
\hline
EncNet+JPU (ours) & ResNet-50 & 51.2\tablefootnote{Following~\cite{zhang2018context}, the mIoU reported in Table~\ref{table:ablation} is on 59 classes w/o background. In this table, the mIoU is measured on 60 classes w/ background for a fair comparison with other methods. Besides, we average the network prediction in multiple scales for evaluation in this table.}\\
EncNet+JPU (ours) & ResNet-101 & \cellcolor{First}\textbf{53.1}\\
\hline
\end{tabular}
\end{center}
	\caption{The state-of-the-art methods on the \textit{val} set of the Pascal Context dataset.}
	\label{table:sota:pascal}
\end{table}
% ---------------------------------------------------------------------------------
\begin{table}
\begin{center}
\begin{tabular}{r|l|cc}
\hline
Method & Backbone & pixAcc\% & mIoU\%\\
\hline
FCN~\cite{long2015fully} & & 71.32 & 29.39\\
SegNet~\cite{badrinarayanan2015segnet} & & 71.00 & 21.64\\
DilatedNet~\cite{Yu2016MultiScaleCA} & & 73.55 & 32.31\\
CascadeNet~\cite{zhou2017scene} & & 74.52 & 34.90\\
RefineNet~\cite{lin2017refinenet} & ResNet-152 & - & 40.7\\
\hline
\multirow{2}{*}{PSPNet~\cite{zhao2017pyramid}} & ResNet-101 & 81.39 & 43.29\\
 & ResNet-269 & 81.69 & 44.94\\
\hline
\multirow{2}{*}{EncNet~\cite{zhang2018context}} & \cellcolor{Second}ResNet-50 & \cellcolor{Second}79.73 & \cellcolor{Second}41.11\\
 & ResNet-101 & 81.69 & 44.65\\
\hline
\hline
\multirow{2}{*}{Ours} & \cellcolor{First}ResNet-50 & \cellcolor{First}80.39 & \cellcolor{First}42.75\\
 & ResNet-101 & 80.99 & 44.34\\
\hline
\end{tabular}
\end{center}
	\caption{Results on the \textit{val} set of ADE20K dataset.}
	\label{table:sota:adeval}
\end{table}
% ---------------------------------------------------------------------------------
\begin{table}
\begin{center}
\begin{tabular}{clc|c}
\hline
Rank & Team & Single Model & Final Score\\
\hline
1 & CASIA\_IVA\_JD & \ding{55} & 0.5547\\
2 & WinterIsComing & \ding{55} & 0.5544\\
- & PSPNet~\cite{zhao2017pyramid} & ResNet-269 & 0.5538\\
- & EncNet~\cite{zhang2018context} & ResNet-101 & \cellcolor{Second}0.5567\\
\hline
- & Ours & ResNet-101 & \cellcolor{First}\textbf{0.5584}\\
\hline
\end{tabular}
\end{center}
	\caption{Results on ADE20K \textit{test} set. The first two entries ranked 1st and 2nd place in COCO-Place challenge 2017.}
	\label{table:sota:adetest}
\end{table}
% ---------------------------------------------------------------------------------
\paragraph{Pascal Context}
In Table~\ref{table:ablation}, our method employs ResNet-50 as the backbone without multi-scale evaluation, and the metrics are calculated on 59 classes excluding background by following~\cite{zhang2018context}.
To compare fairly with the state-of-the-art methods, we average the prediction in multiple scales and calculate the metrics among 60 classes including background, which are then reported in Table~\ref{table:sota:pascal}.
With ResNet-50 as the backbone, our method outperforms DeepLabV2 (with COCO pretraining) and RefineNet by a large margin, which employ ResNet-101 and ResNet-152 as the backbone, respectively.
Moreover, our method (ResNet-50) achieves competitive performance compared to EncNet with ResNet-101 as the backbone.
By replacing ResNet-50 with a deeper network ResNet-101, our method gets an additional 1.9\% improvement in mIoU, which outperforms EncNet (ResNet-101) and DUpsampling (Xception-71) significantly and achieves the state-of-the-art performance.
Notably, Xception-71 is a much stronger backbone than ResNet-101.
For completeness, we also report the mIoU on 59 classes (w/o background), which is 52.10\%(ResNet-50) and 54.03\% (ResNet-101).
% ---------------------------------------------------------------------------------
\vspace{-1em}
\paragraph{ADE20K}
ADE20K dataset~\cite{zhou2017scene} is a scene parsing benchmark, which contains 150 stuff/object categories.
The dataset includes 20K/2K/3K images for training (\textit{train}), validation (\textit{val}), and testing (\textit{test}).

We train our network on the \textit{train} set for 120 epochs with learning rate 0.01.
We then evaluate the model on the \textit{val} set and report pixAcc and mIoU in Table~\ref{table:sota:adeval}.
When employing ResNet-50 as the backbone, our method outperforms EncNet (ResNet-50) by 1.64\% in mIoU, while achieving a much better performance compared to RefineNet (ResNet-152).
By replacing ResNet-50 with ResNet-101, our method obtains competitive performance compared to EncNet (ResNet-101) and PSPNet (ResNet-269).
Our method (ResNet-101) performs a little worse than EncNet, and we attribute this to the spatial resolution of the training images.
Concretely, in our method, the training images are cropped to $480\times 480$ for processing 4 images in a GPU with 12G memory.
However, EncNet is trained with $576\times 576$ images on GPUs with memory larger than 12G.

We then fine-tune our network on the \textit{train} set and \textit{val} set for another 20 epochs with learning rate 0.001.
The predictions on the \textit{test} set are submitted to the evaluation server.
As shown in Table~\ref{table:sota:adetest}, our method outperforms two winning entries from the COCO-Place challenge 2017.
Moreover, our method also achieves better performance compared to PSPNet and EncNet, although it performs worse on the \textit{val} set.
Notably, Final Score is the metric used in the evaluation server, which is the average of pixAcc and mIoU.

The visual results from both the Pascal Context dataset and the ADE20K dataset are shown in Figure~\ref{fig:resnet101}.
More results are shown in the supplementary material.
% ---------------------------------------------------------------------------------
\begin{figure} 
\begin{center}
	\begin{subfigure}[b]{0.24\linewidth}
		\includegraphics[width=\linewidth]{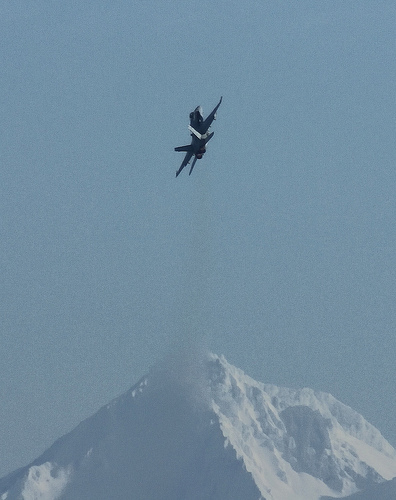}\\
		\includegraphics[width=\linewidth]{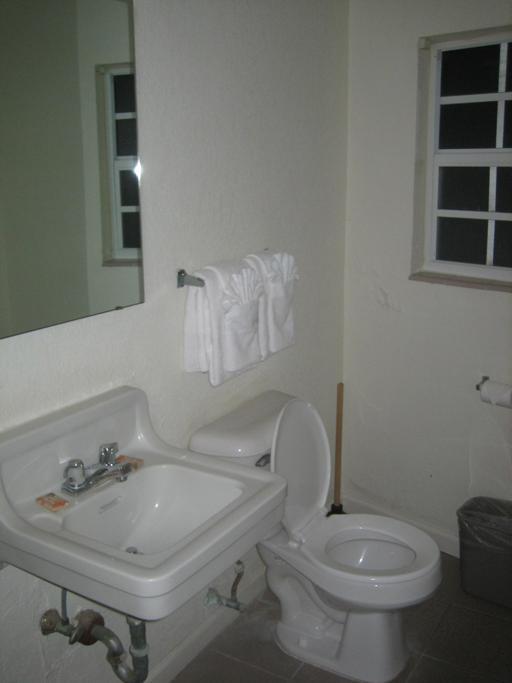}
      	\caption{Input}
      	\label{fig:resnet101:input}
    \end{subfigure}
    \hfill
    \begin{subfigure}[b]{0.24\linewidth}
		\includegraphics[width=\linewidth]{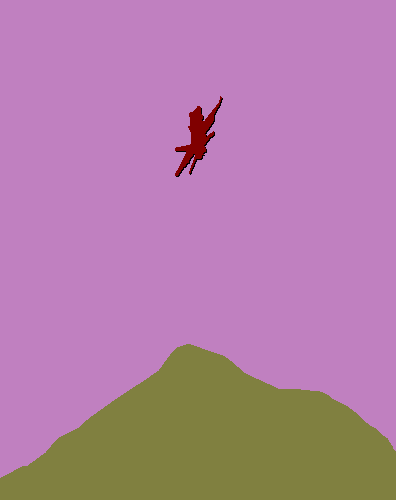}\\
		\includegraphics[width=\linewidth]{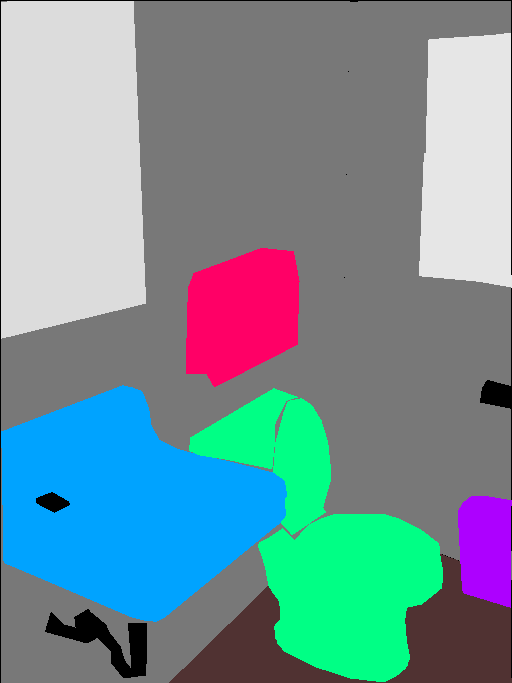}
      	\caption{GT}
      	\label{fig:resnet101:gt}
    \end{subfigure}
    \hfill
	\begin{subfigure}[b]{0.24\linewidth}
		\includegraphics[width=\linewidth]{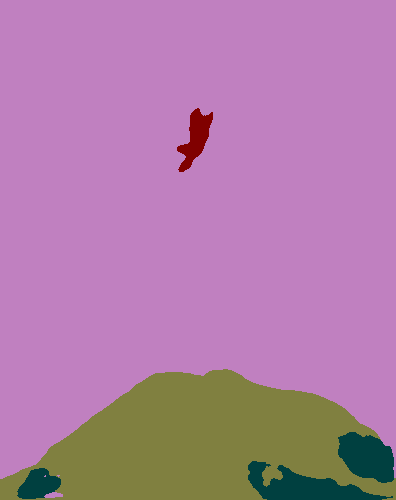}\\
		\includegraphics[width=\linewidth]{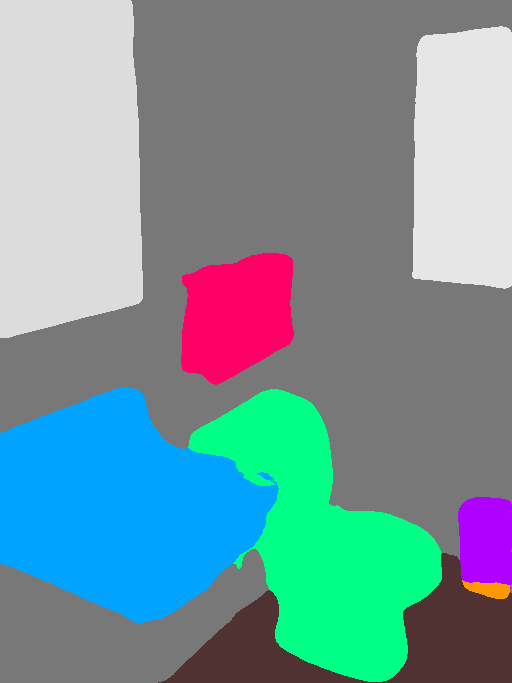}
      	\caption{EncNet~\cite{zhang2018context}}
      	\label{fig:resnet101:encnet}
    \end{subfigure}
    \hfill
   	\begin{subfigure}[b]{0.24\linewidth}
		\includegraphics[width=\linewidth]{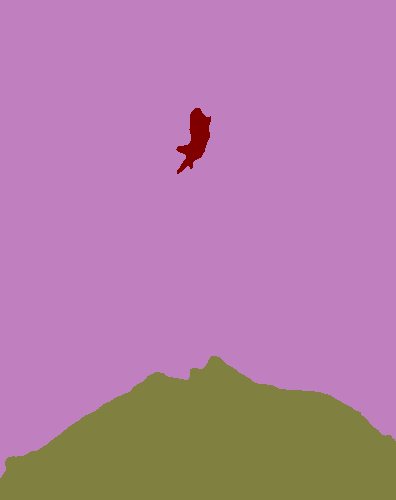}\\
		\includegraphics[width=\linewidth]{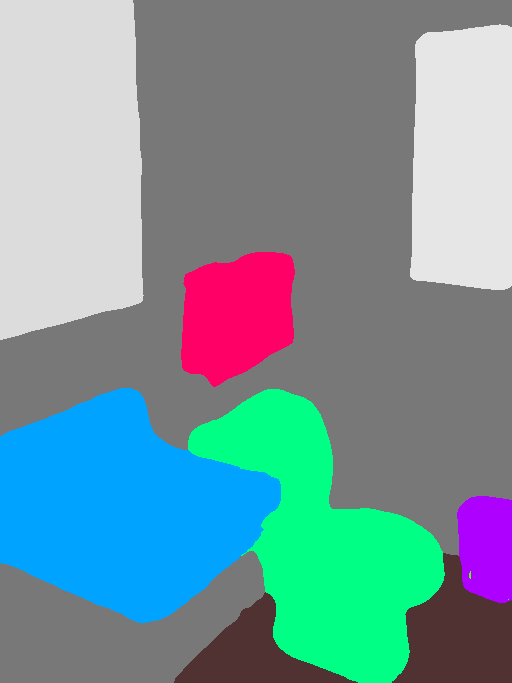}
      	\caption{Ours}
      	\label{fig:resnet101:ours}
    \end{subfigure}
\end{center}
	\caption{Visual results of our method (ResNet-101). The first row is from Pascal Context \textit{val} set, while the second row is from ADE20K \textit{val} set. Best viewed in color.}
	\label{fig:resnet101}
\end{figure}
% ---------------------------------------------------------------------------------
\section{Conclusion}
In this paper, we have analyzed the differences and connections between dilated convolution and stride convolution.
Based on the analysis, we formulated the task of extracting high-resolution feature maps into a joint upsampling problem and proposed a novel CNN module JPU to solve the problem.
By replacing the time and memory consuming dilated convolutions with our JPU, the computation complexity is reduced by more than three times without performance loss.
The ablation study shows that the proposed JPU is superior to other upsampling modules.
By plugging JPU, several modern approaches for semantic segmentation achieve a better performance while runs much faster than before.
Results on two segmentation datasets show that our method achieves the state-of-the-art performance while reducing the computation complexity dramatically. 
% ---------------------------------------------------------------------------------

{\small
\bibliographystyle{ieee}
\bibliography{egbib}
}

\section*{More Visual Results}
\begin{figure*}[!htb] 
\begin{center}
	\begin{subfigure}[b]{0.24\linewidth}
		\includegraphics[width=\linewidth]{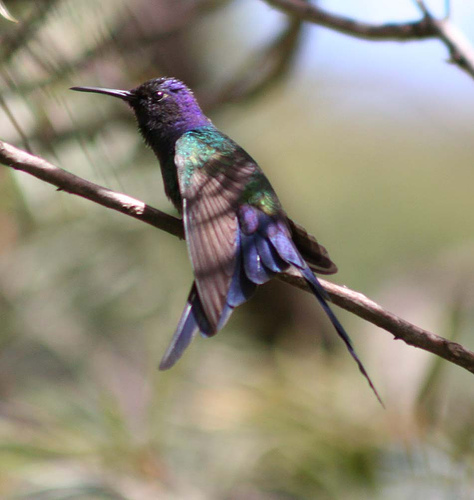}\\
		\includegraphics[width=\linewidth]{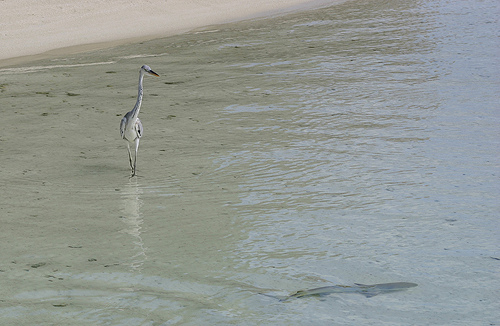}\\
		\includegraphics[width=\linewidth]{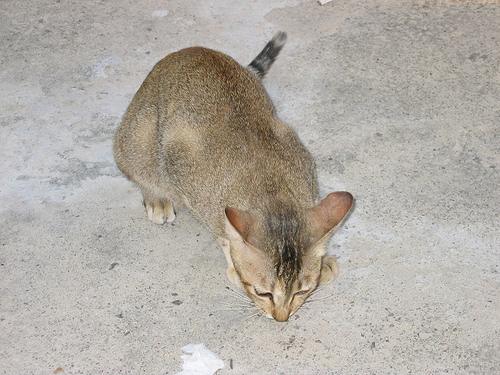}\\
		\includegraphics[width=\linewidth]{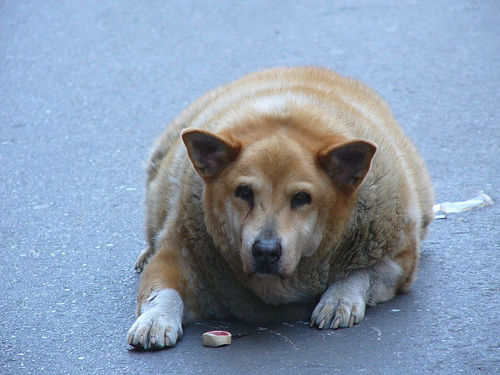}\\
		\includegraphics[width=\linewidth]{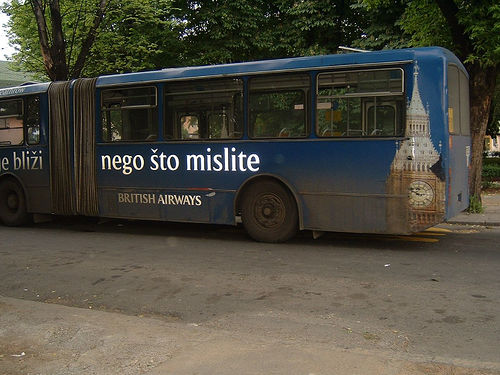}\\
		\includegraphics[width=\linewidth]{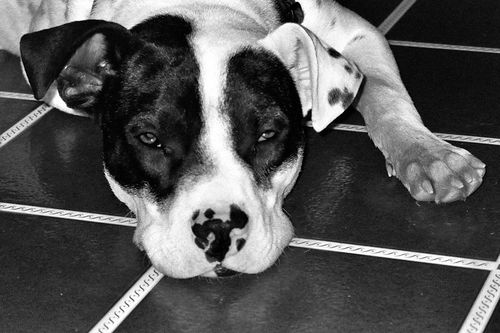}\\
      	\caption{Input}
    \end{subfigure}
    \hfill
	\begin{subfigure}[b]{0.24\linewidth}
		\includegraphics[width=\linewidth]{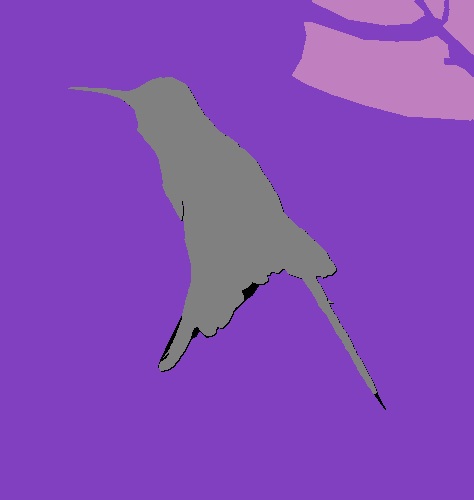}\\
		\includegraphics[width=\linewidth]{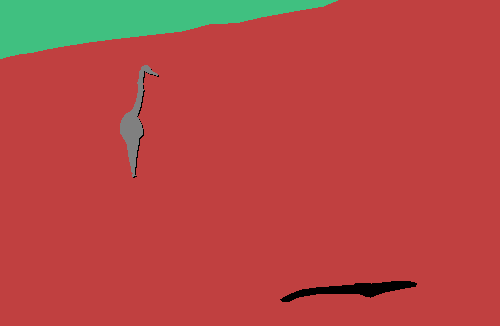}\\
		\includegraphics[width=\linewidth]{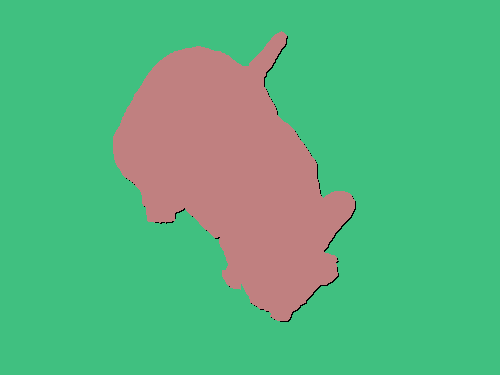}\\
		\includegraphics[width=\linewidth]{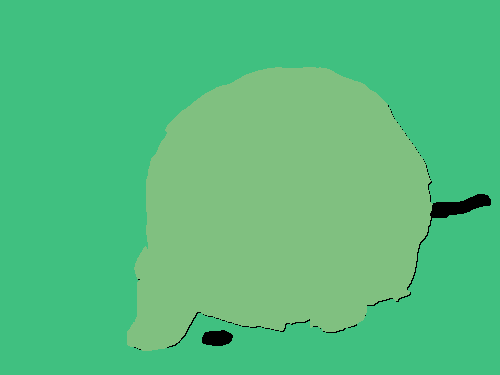}\\
		\includegraphics[width=\linewidth]{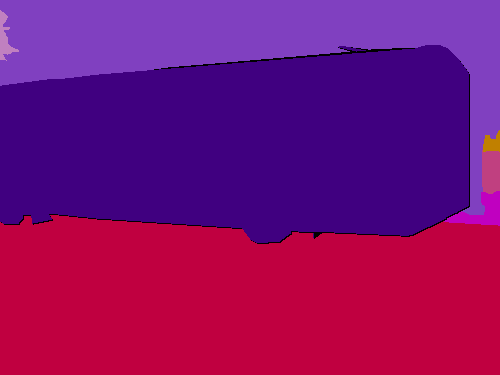}\\
		\includegraphics[width=\linewidth]{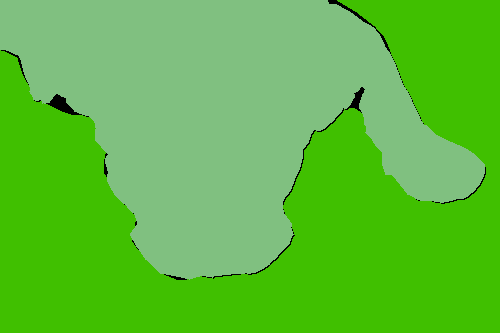}\\
      	\caption{GT}
    \end{subfigure}
    \hfill
	\begin{subfigure}[b]{0.24\linewidth}
		\includegraphics[width=\linewidth]{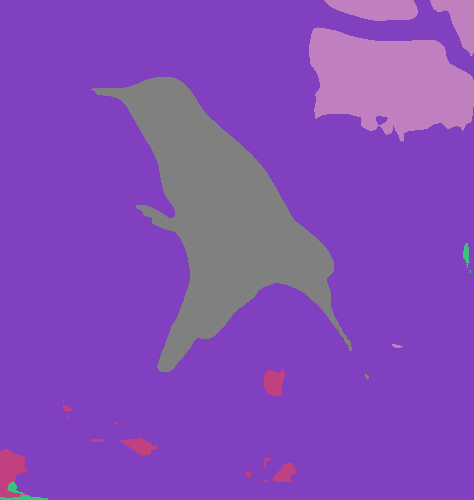}\\
		\includegraphics[width=\linewidth]{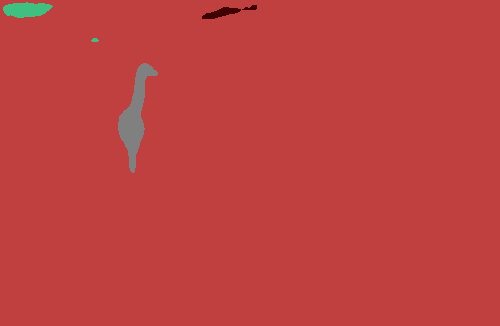}\\
		\includegraphics[width=\linewidth]{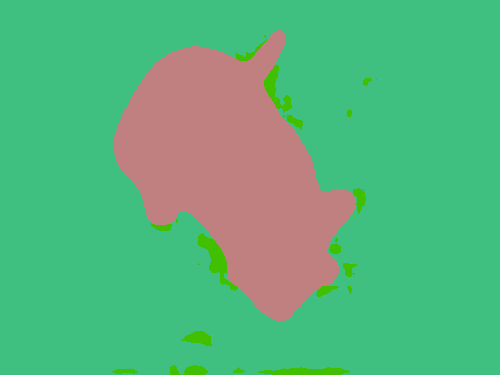}\\
		\includegraphics[width=\linewidth]{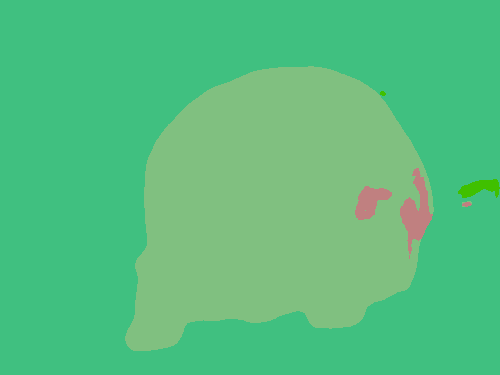}\\
		\includegraphics[width=\linewidth]{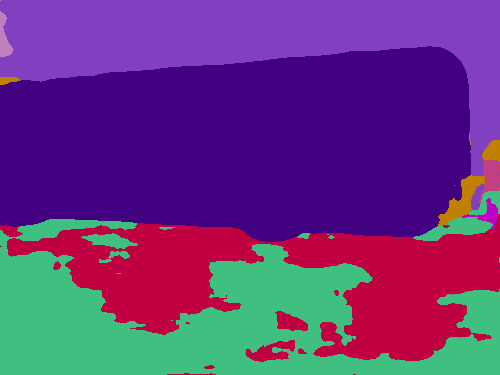}\\
		\includegraphics[width=\linewidth]{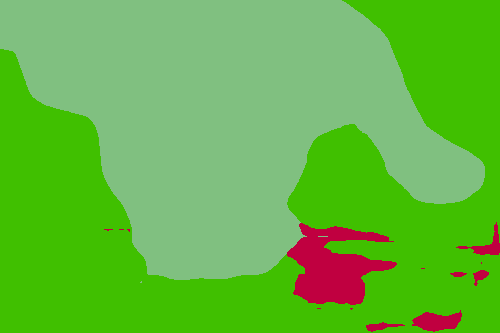}\\
      	\caption{EncNet}
    \end{subfigure}
    \hfill
    \begin{subfigure}[b]{0.24\linewidth}
		\includegraphics[width=\linewidth]{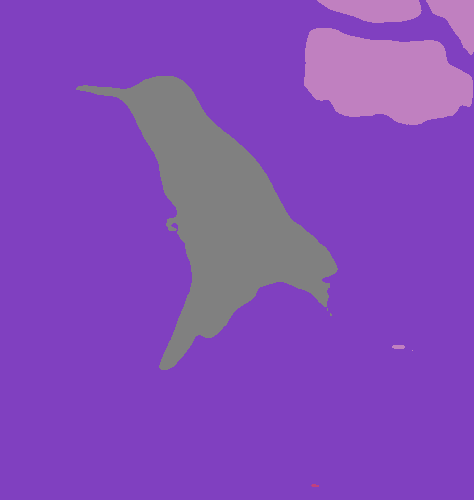}\\
		\includegraphics[width=\linewidth]{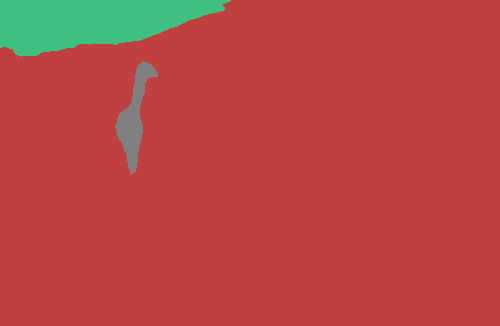}\\
		\includegraphics[width=\linewidth]{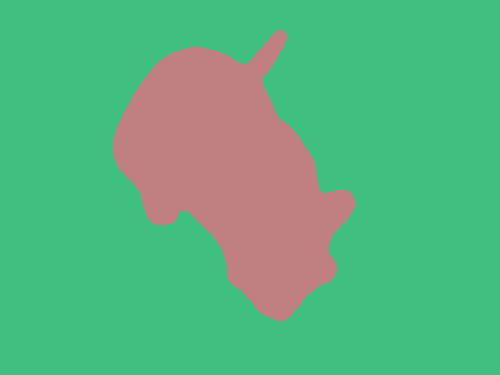}\\
		\includegraphics[width=\linewidth]{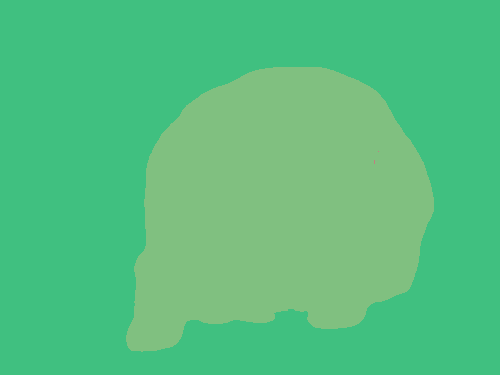}\\
		\includegraphics[width=\linewidth]{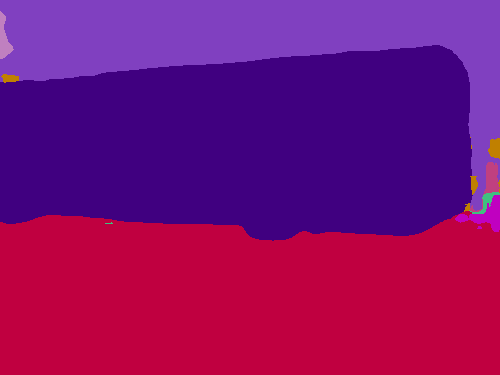}\\
		\includegraphics[width=\linewidth]{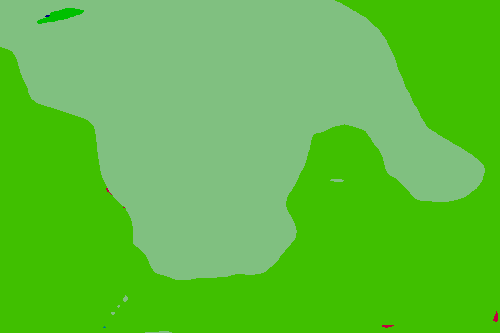}\\
      	\caption{Ours}
    \end{subfigure}    
\end{center}
	\caption{Visual results of our method (ResNet-101) on the Pascal Context \textit{val} set. Best viewed in color.}
	\label{img:visual_first}
\end{figure*}

\begin{figure*}[!htb] 
\begin{center}
	\begin{subfigure}[b]{0.24\linewidth}
		\includegraphics[width=\linewidth]{images/supplementary/img_ADE_val_00000767}\\
		\includegraphics[width=\linewidth]{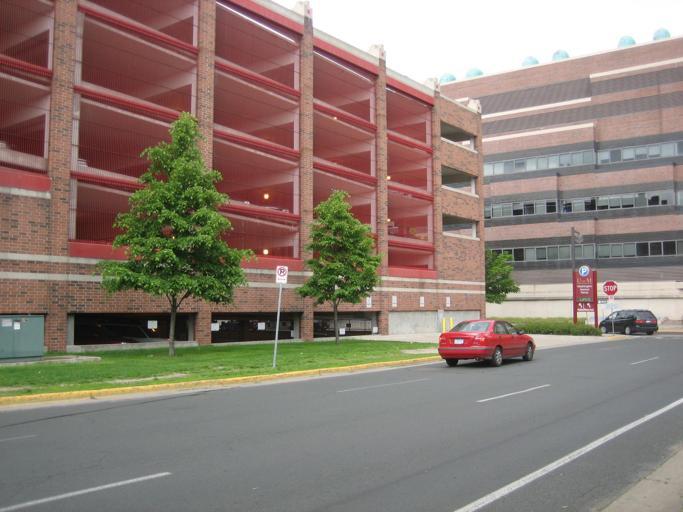}\\
		\includegraphics[width=\linewidth]{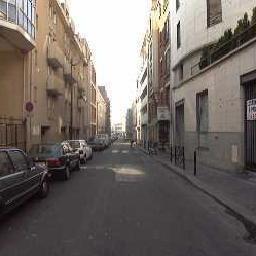}\\
		\includegraphics[width=\linewidth]{images/supplementary/img_ADE_val_00000925}\\
		\includegraphics[width=\linewidth]{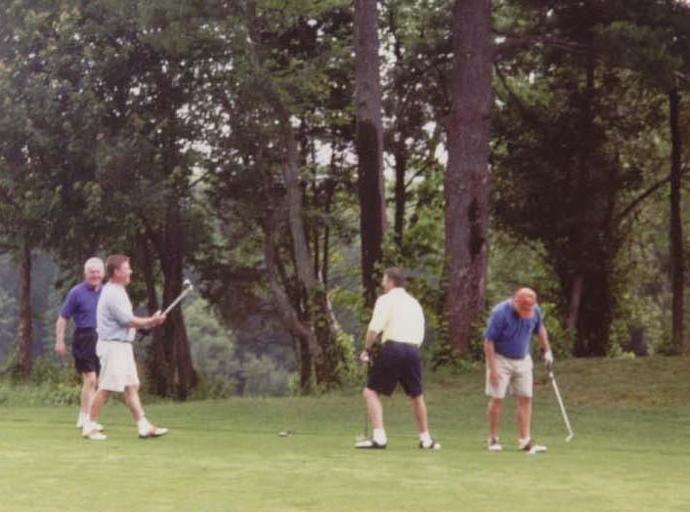}\\
		\caption{Input}
    \end{subfigure}
    \hfill
    \begin{subfigure}[b]{0.24\linewidth}
		\includegraphics[width=\linewidth]{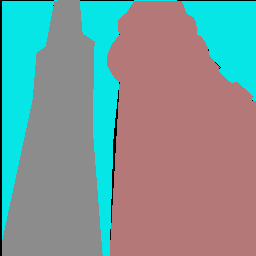}\\
		\includegraphics[width=\linewidth]{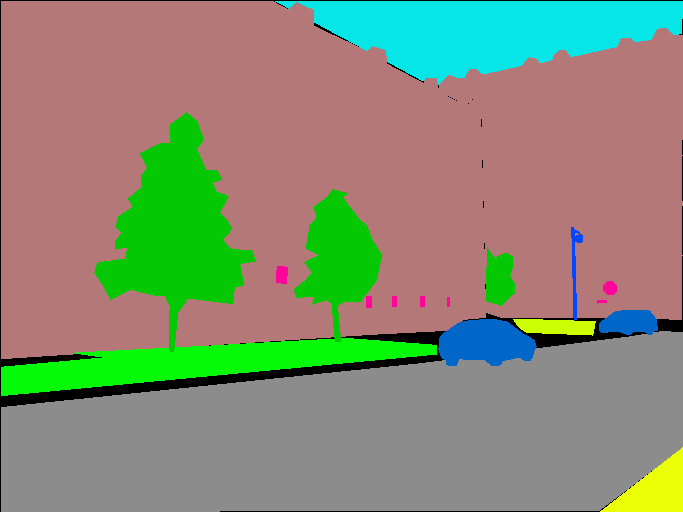}\\
		\includegraphics[width=\linewidth]{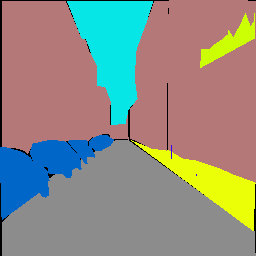}\\
		\includegraphics[width=\linewidth]{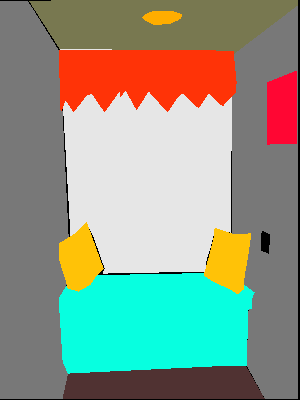}\\
		\includegraphics[width=\linewidth]{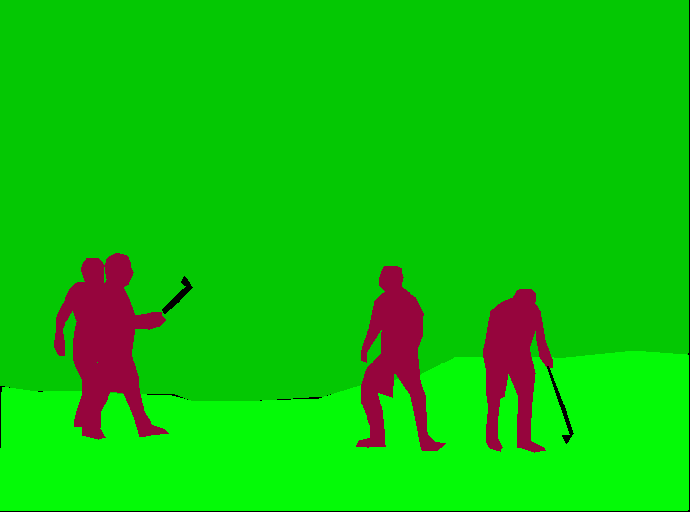}\\
		\caption{GT}
    \end{subfigure}
    \hfill
    \begin{subfigure}[b]{0.24\linewidth}
		\includegraphics[width=\linewidth]{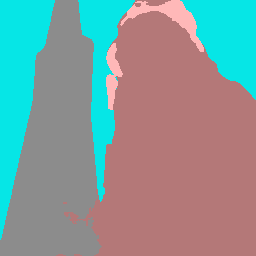}\\
		\includegraphics[width=\linewidth]{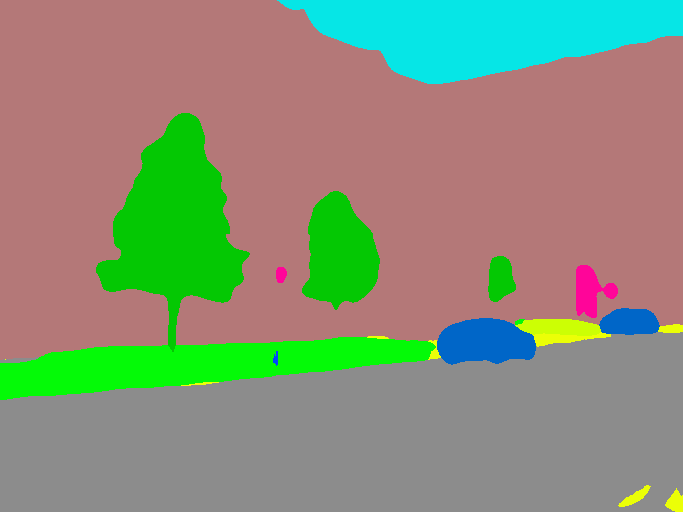}\\
		\includegraphics[width=\linewidth]{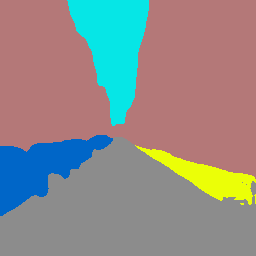}\\
		\includegraphics[width=\linewidth]{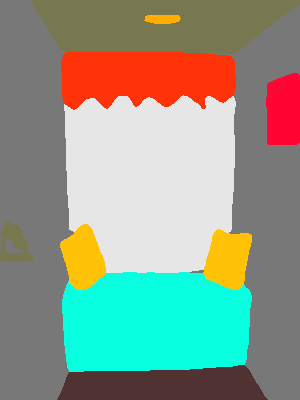}\\
		\includegraphics[width=\linewidth]{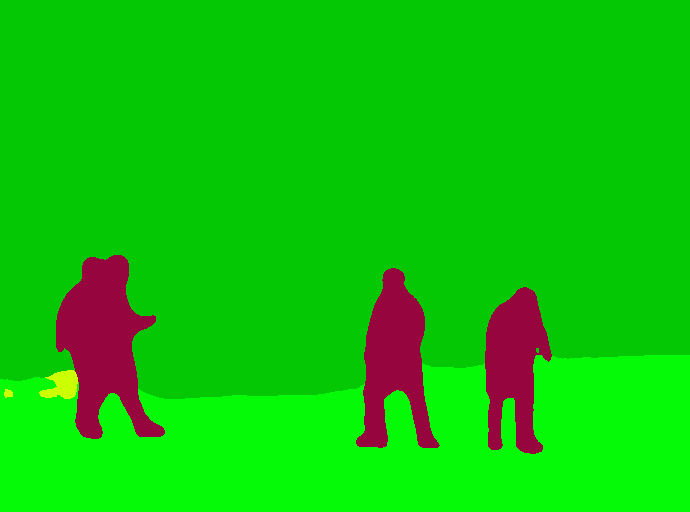}\\
		\caption{EncNet}
    \end{subfigure}
    \hfill
    \begin{subfigure}[b]{0.24\linewidth}
		\includegraphics[width=\linewidth]{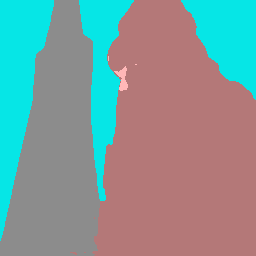}\\
		\includegraphics[width=\linewidth]{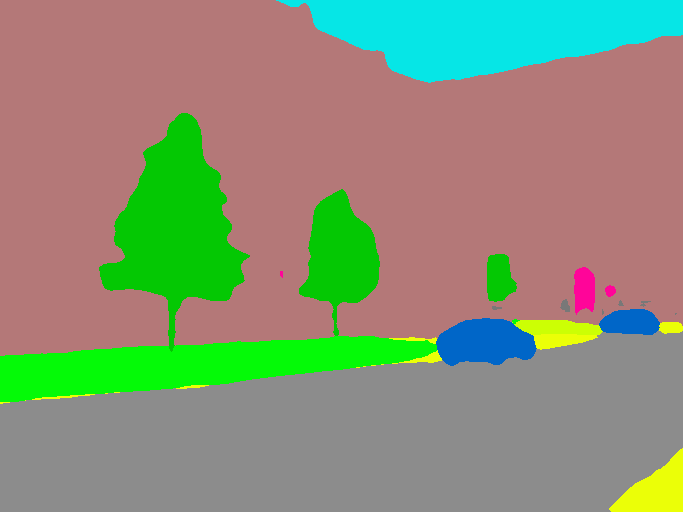}\\
		\includegraphics[width=\linewidth]{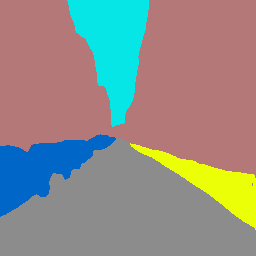}\\
		\includegraphics[width=\linewidth]{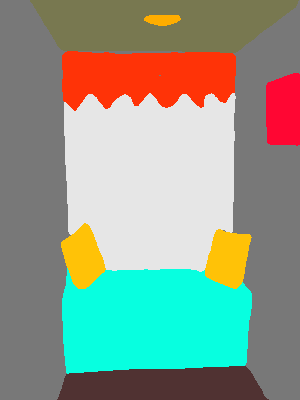}\\
		\includegraphics[width=\linewidth]{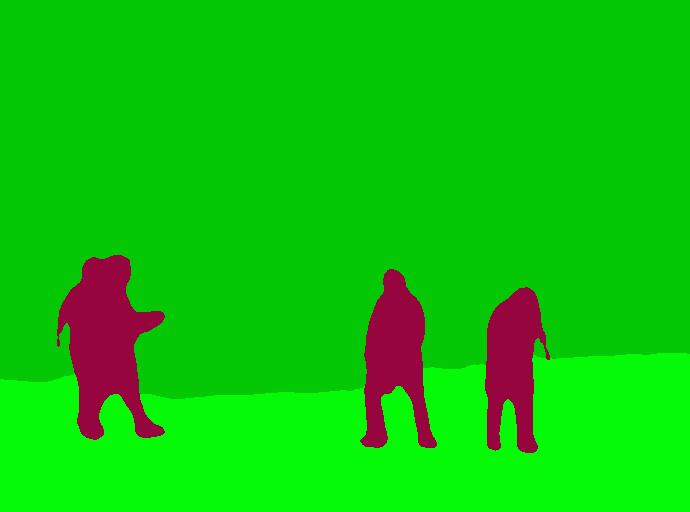}\\
		\caption{Ours}
    \end{subfigure}
\end{center}
	\caption{Visual results of our method (ResNet-101) on the ADE20K \textit{val} set. Best viewed in color.}
	\label{img:visual_last}
\end{figure*}

See Figure~\ref{img:visual_first}$-$\ref{img:visual_last} in the following pages for more visual results.

\begin{figure*}[!htb] 
\begin{center}
	\begin{subfigure}[b]{0.24\linewidth}
		\includegraphics[width=\linewidth]{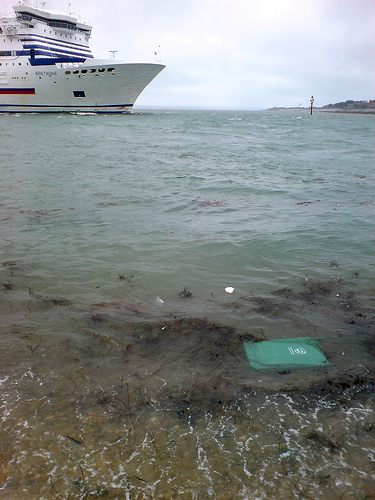}\\
		\includegraphics[width=\linewidth]{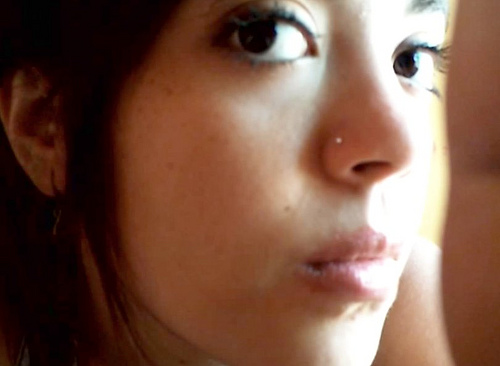}\\
		\includegraphics[width=\linewidth]{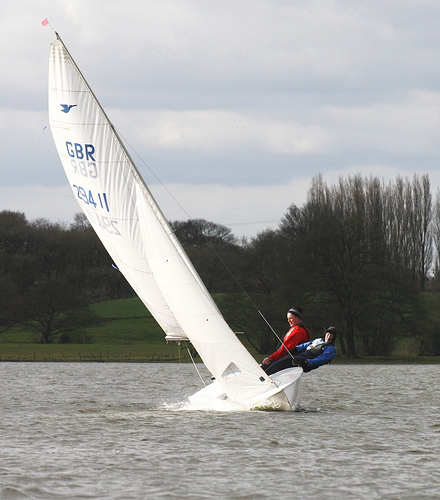}\\
		\includegraphics[width=\linewidth]{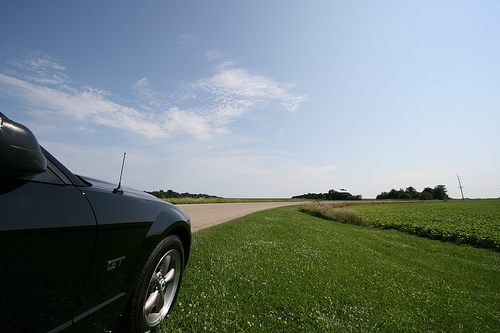}\\
		\includegraphics[width=\linewidth]{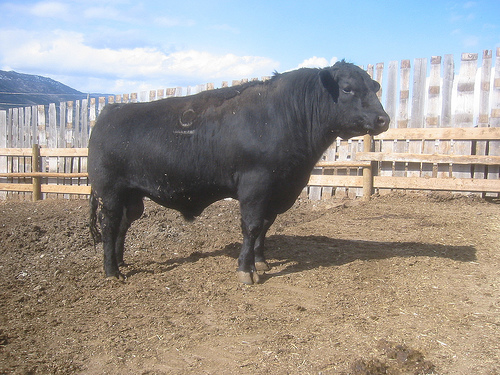}\\
		\caption{Input}
    \end{subfigure}
    \hfill
	\begin{subfigure}[b]{0.24\linewidth}
		\includegraphics[width=\linewidth]{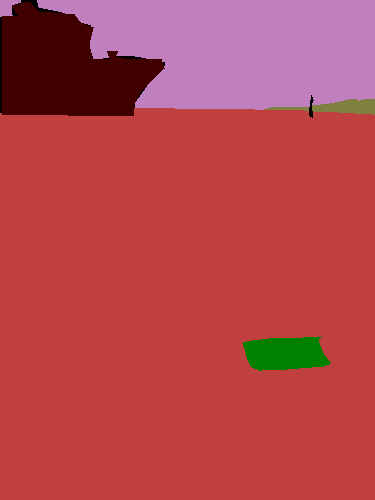}\\
		\includegraphics[width=\linewidth]{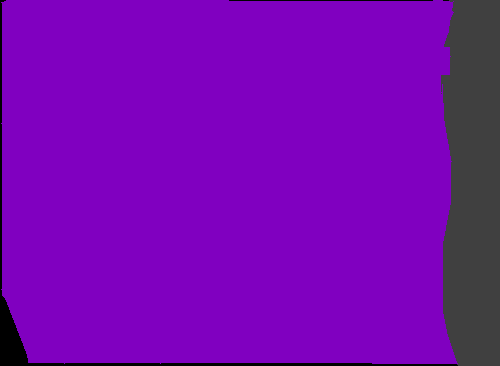}\\
		\includegraphics[width=\linewidth]{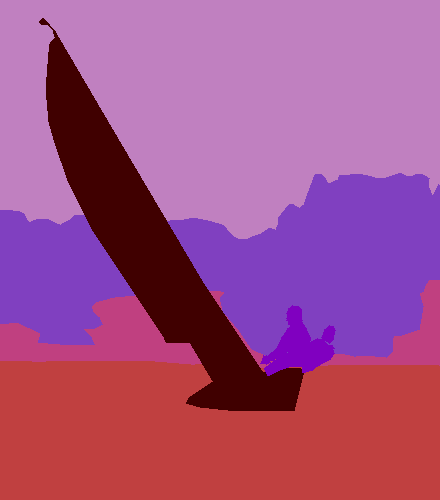}\\
		\includegraphics[width=\linewidth]{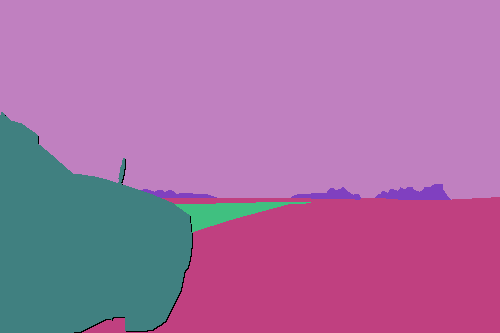}\\
		\includegraphics[width=\linewidth]{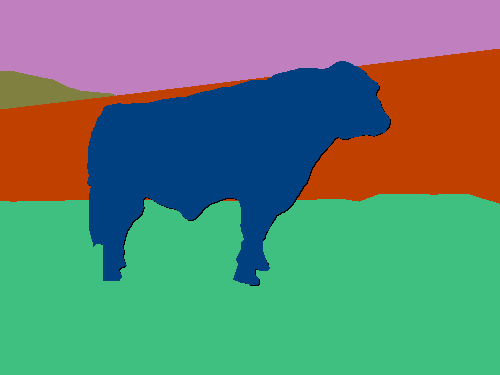}\\
		\caption{GT}
    \end{subfigure}
    \hfill
	\begin{subfigure}[b]{0.24\linewidth}
		\includegraphics[width=\linewidth]{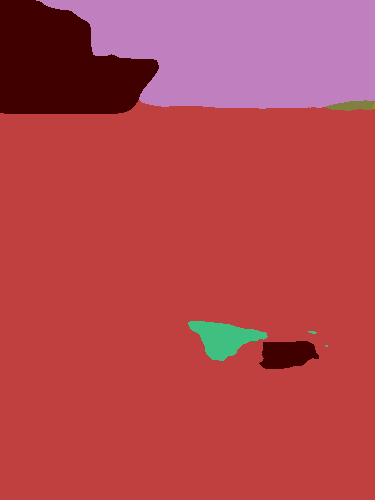}\\
		\includegraphics[width=\linewidth]{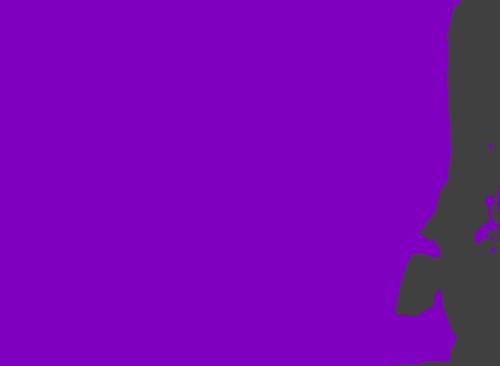}\\
		\includegraphics[width=\linewidth]{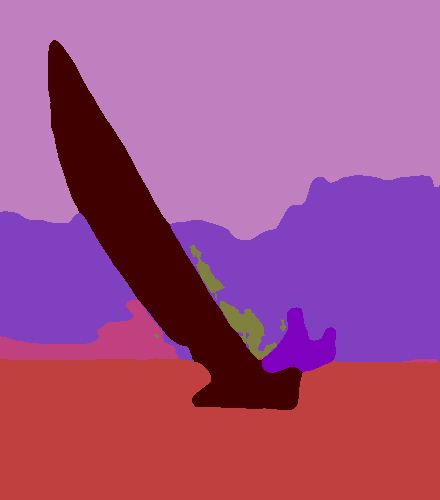}\\
		\includegraphics[width=\linewidth]{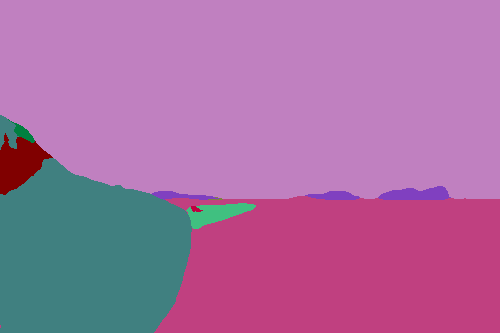}\\
		\includegraphics[width=\linewidth]{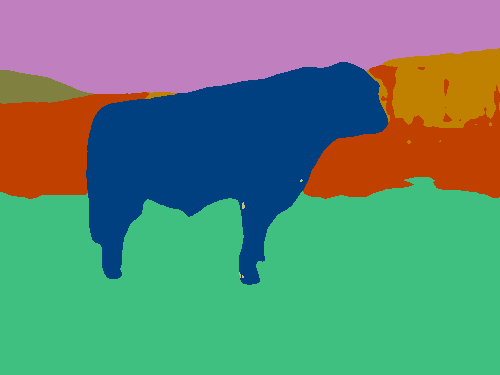}\\
		\caption{EncNet}
    \end{subfigure}
    \hfill
	\begin{subfigure}[b]{0.24\linewidth}
		\includegraphics[width=\linewidth]{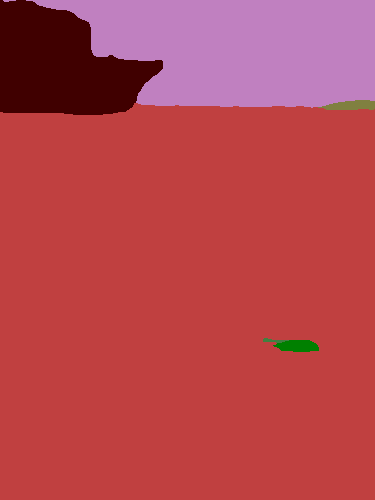}\\
		\includegraphics[width=\linewidth]{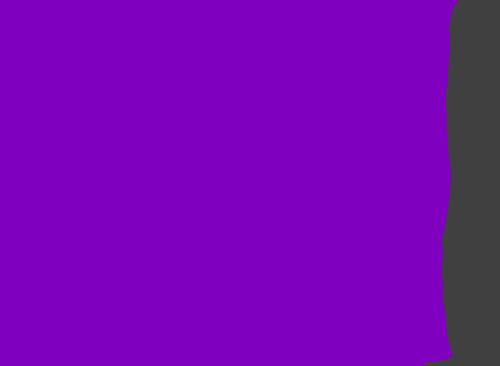}\\
		\includegraphics[width=\linewidth]{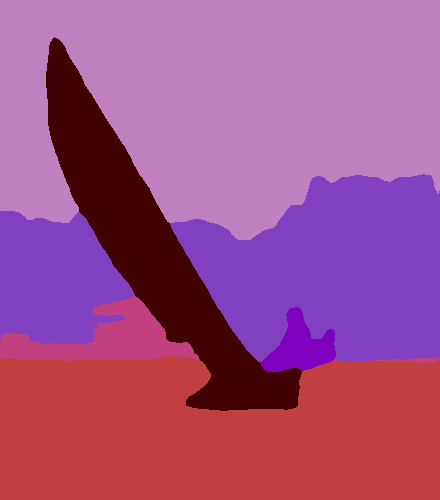}\\
		\includegraphics[width=\linewidth]{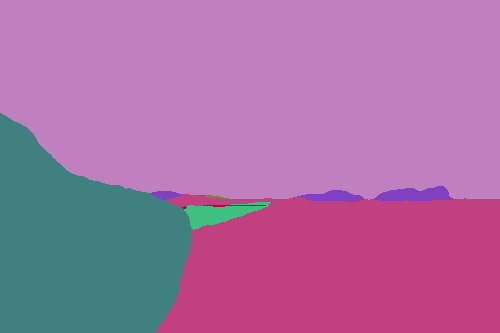}\\
		\includegraphics[width=\linewidth]{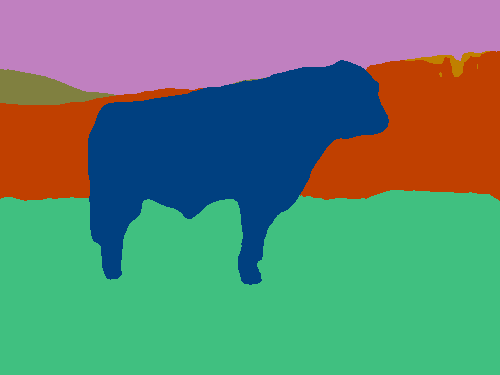}\\
		\caption{Ours}
    \end{subfigure}
\end{center}
	\caption{Visual results of our method (ResNet-101) on the Pascal Context \textit{val} set. Best viewed in color.}
\end{figure*}

\begin{figure*}[!htb] 
\begin{center}
	\begin{subfigure}[b]{0.24\linewidth}
		\includegraphics[width=\linewidth]{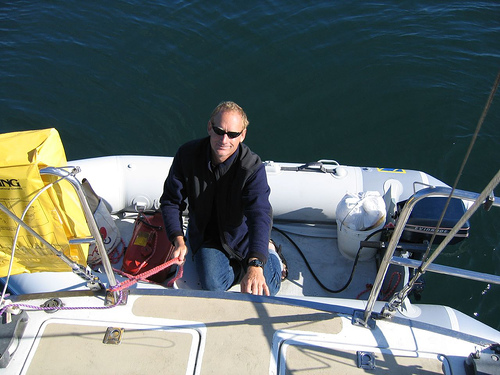}\\
		\includegraphics[width=\linewidth]{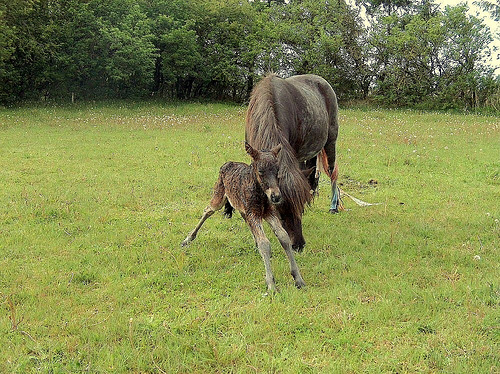}\\
		\includegraphics[width=\linewidth]{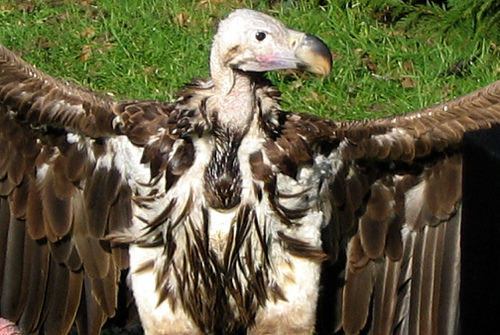}\\
		\includegraphics[width=\linewidth]{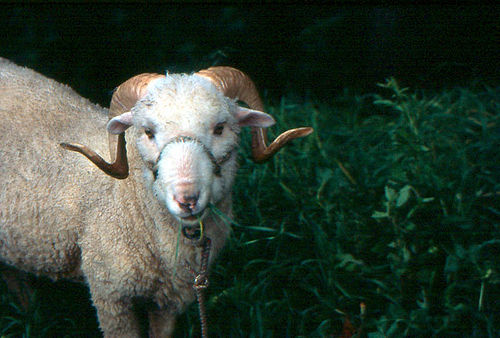}\\
		\includegraphics[width=\linewidth]{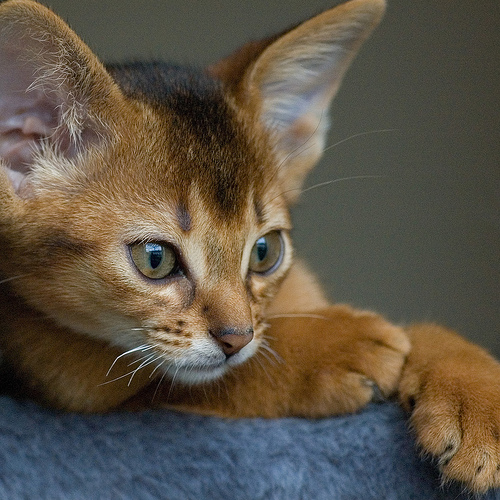}\\
		\includegraphics[width=\linewidth]{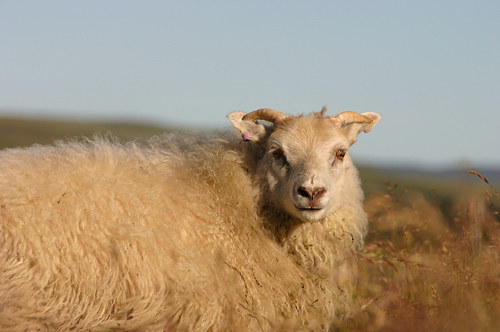}\\
		\caption{Input}
    \end{subfigure}
    \hfill
	\begin{subfigure}[b]{0.24\linewidth}
		\includegraphics[width=\linewidth]{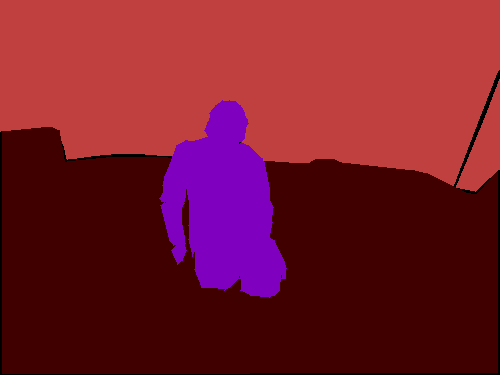}\\
		\includegraphics[width=\linewidth]{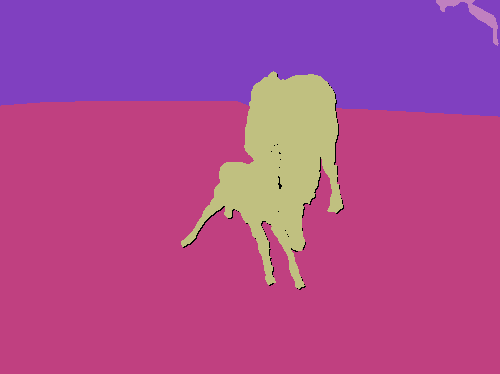}\\
		\includegraphics[width=\linewidth]{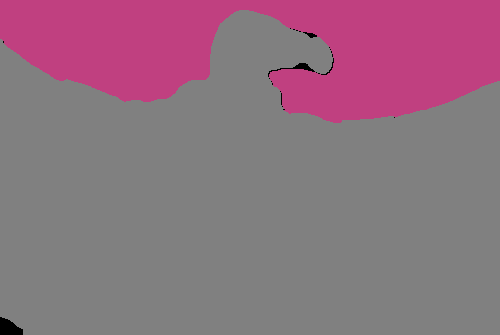}\\
		\includegraphics[width=\linewidth]{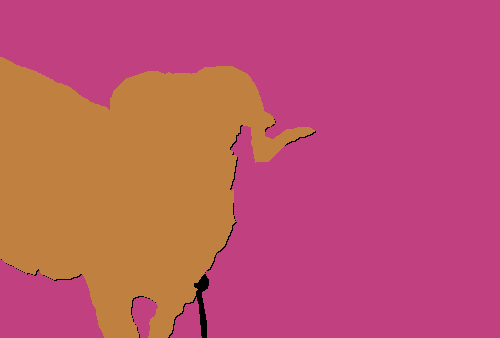}\\
		\includegraphics[width=\linewidth]{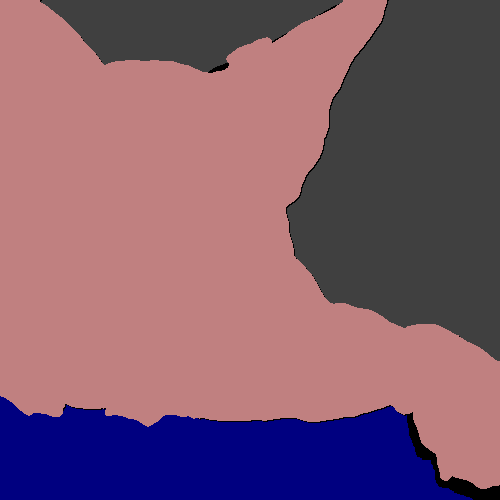}\\
		\includegraphics[width=\linewidth]{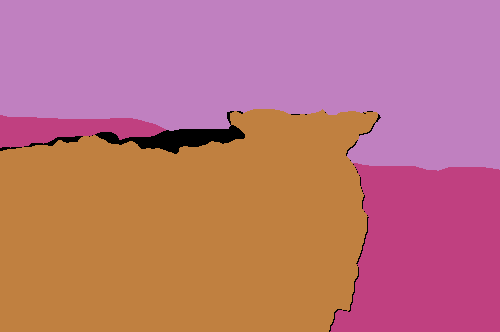}\\
		\caption{GT}
    \end{subfigure}
    \hfill
	\begin{subfigure}[b]{0.24\linewidth}
		\includegraphics[width=\linewidth]{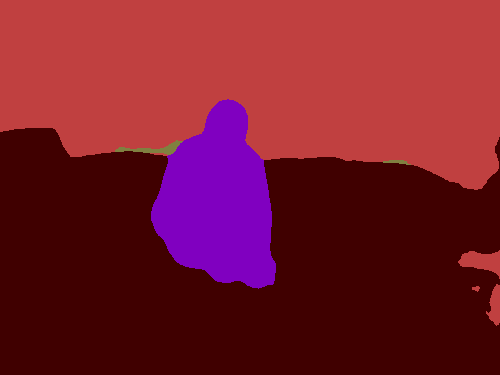}\\
		\includegraphics[width=\linewidth]{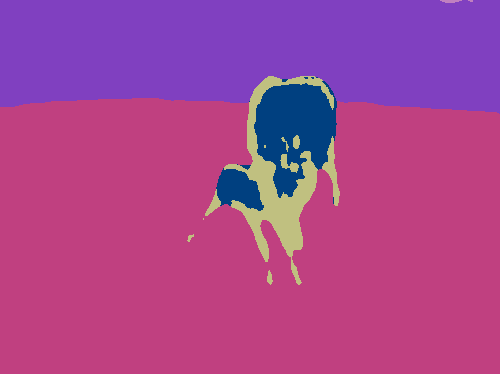}\\
		\includegraphics[width=\linewidth]{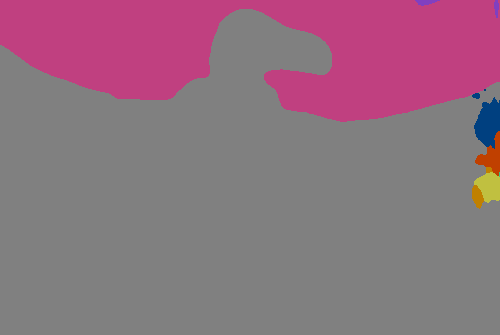}\\
		\includegraphics[width=\linewidth]{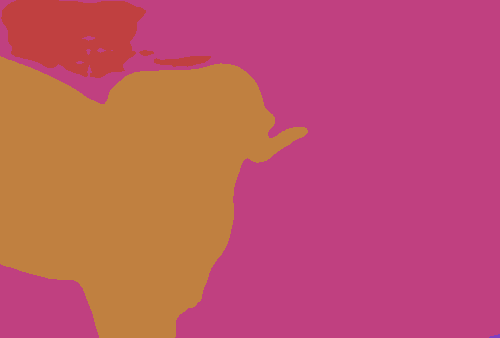}\\
		\includegraphics[width=\linewidth]{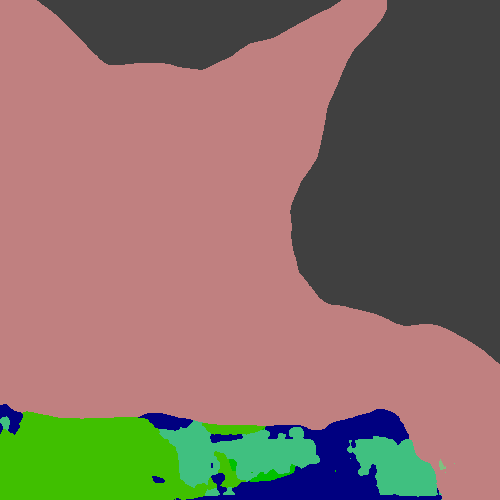}\\
		\includegraphics[width=\linewidth]{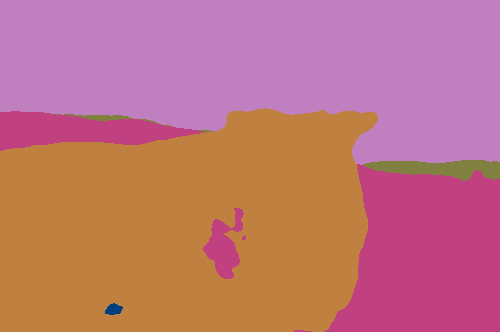}\\
		\caption{EncNet}
    \end{subfigure}
    \hfill
	\begin{subfigure}[b]{0.24\linewidth}
		\includegraphics[width=\linewidth]{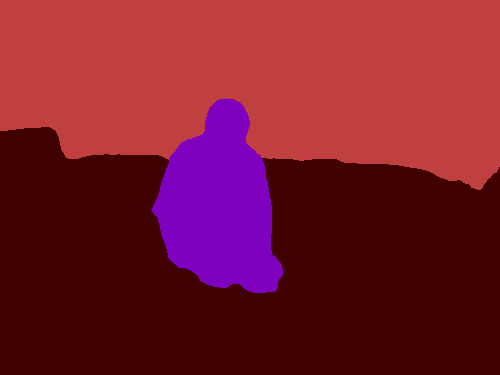}\\
		\includegraphics[width=\linewidth]{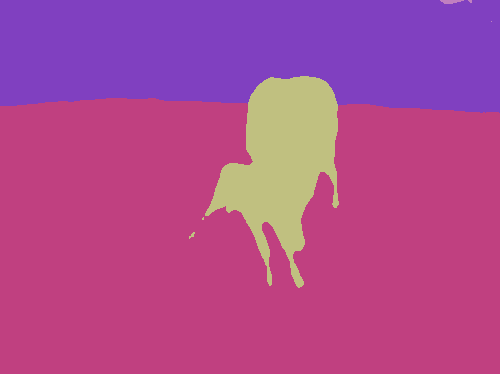}\\
		\includegraphics[width=\linewidth]{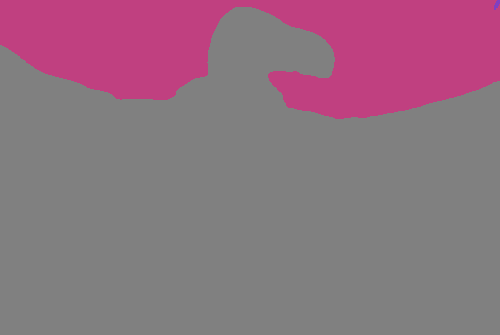}\\
		\includegraphics[width=\linewidth]{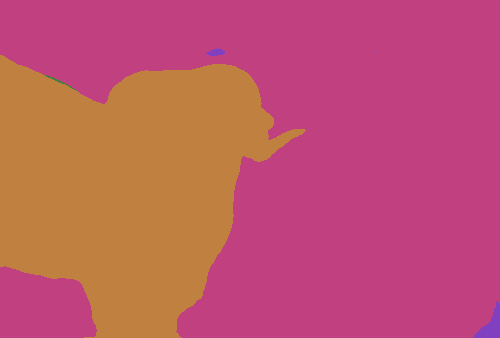}\\
		\includegraphics[width=\linewidth]{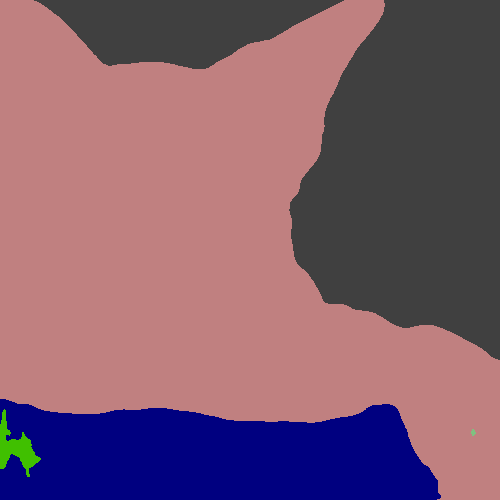}\\
		\includegraphics[width=\linewidth]{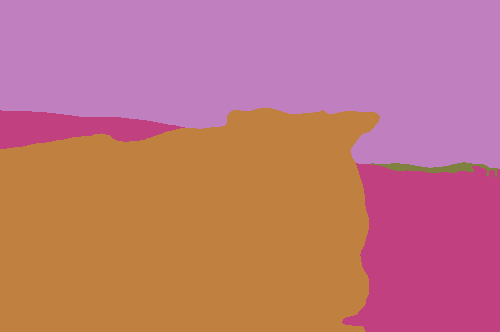}\\
		\caption{Ours}
    \end{subfigure}
\end{center}
	\caption{Visual results of our method (ResNet-101) on the Pascal Context \textit{val} set. Best viewed in color.}
\end{figure*}

\begin{figure*}[!htb] 
\begin{center}
	\begin{subfigure}[b]{0.24\linewidth}
		\includegraphics[width=\linewidth]{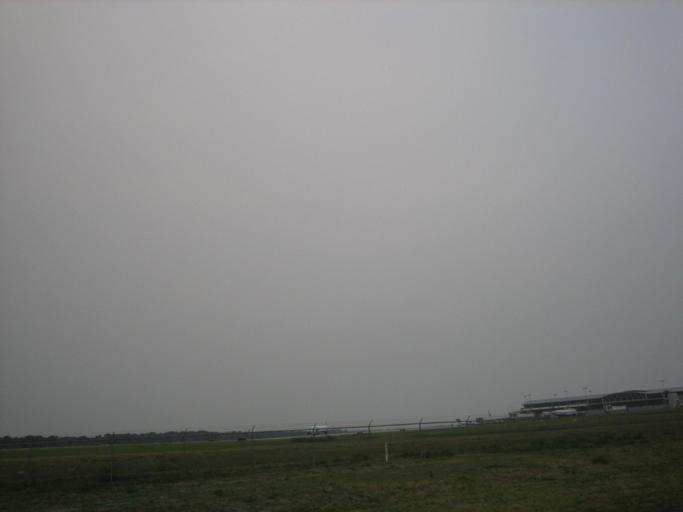}\\
		\includegraphics[width=\linewidth]{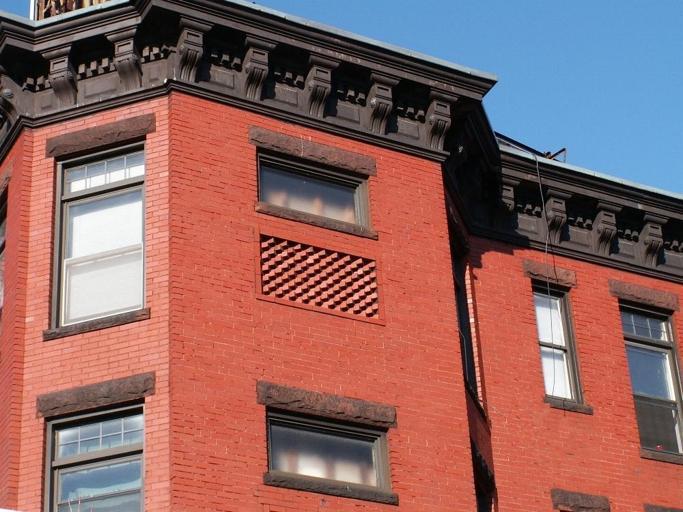}\\
		\includegraphics[width=\linewidth]{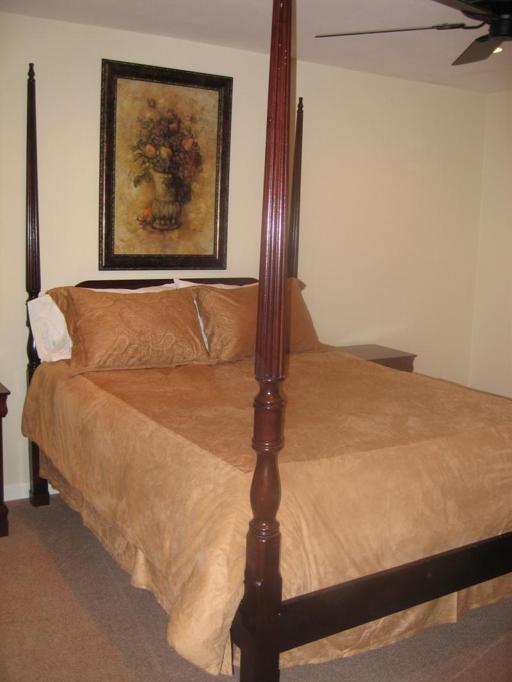}\\
		\includegraphics[width=\linewidth]{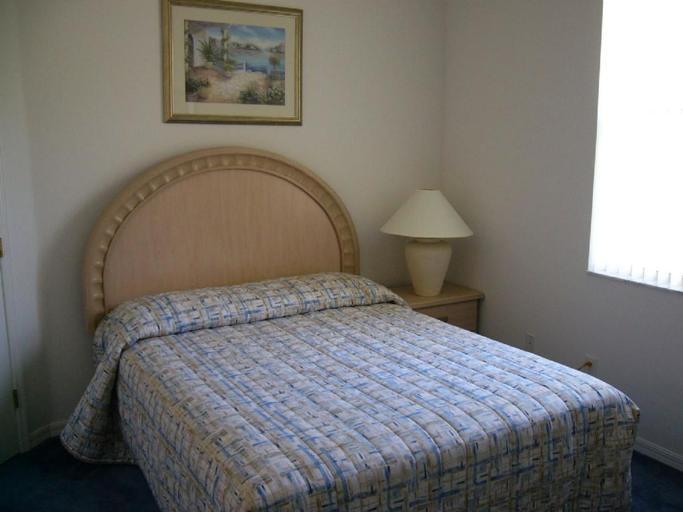}\\
		\includegraphics[width=\linewidth]{images/supplementary/img_ADE_val_00000178}\\
		\includegraphics[width=\linewidth]{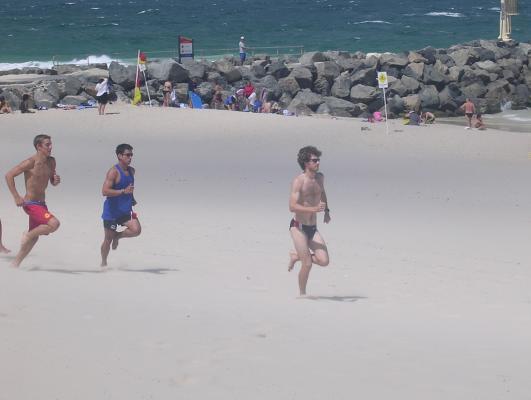}\\
		\caption{Input}
    \end{subfigure}
    \hfill
	\begin{subfigure}[b]{0.24\linewidth}
		\includegraphics[width=\linewidth]{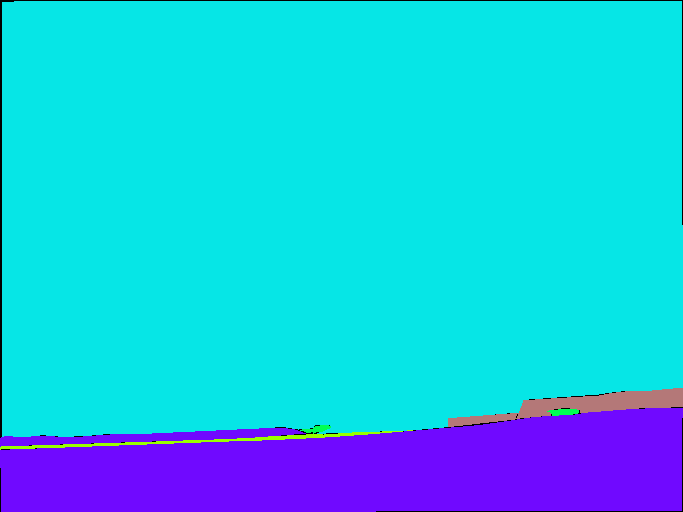}\\
		\includegraphics[width=\linewidth]{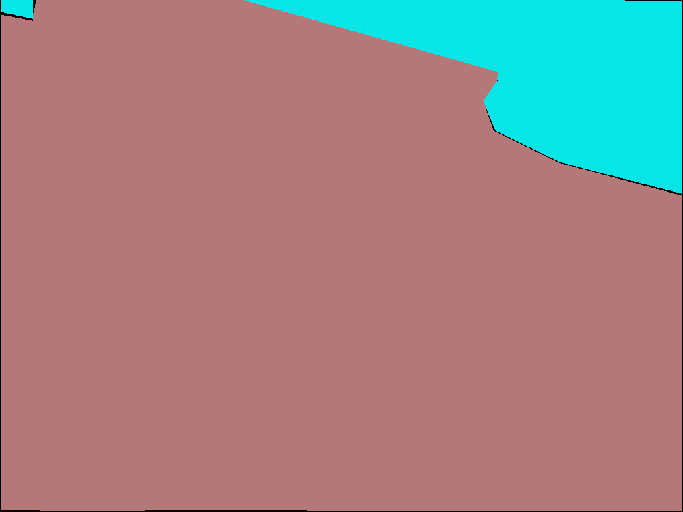}\\
		\includegraphics[width=\linewidth]{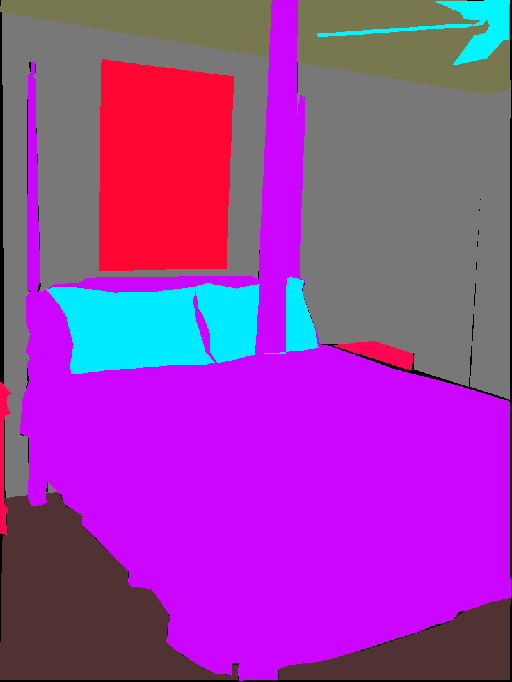}\\
		\includegraphics[width=\linewidth]{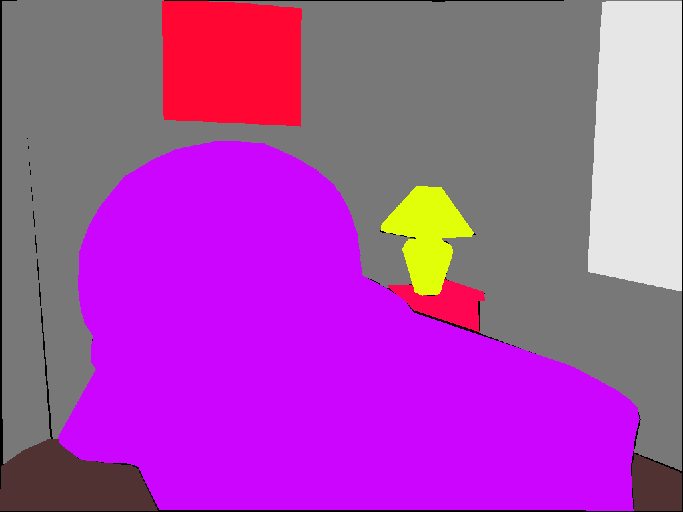}\\
		\includegraphics[width=\linewidth]{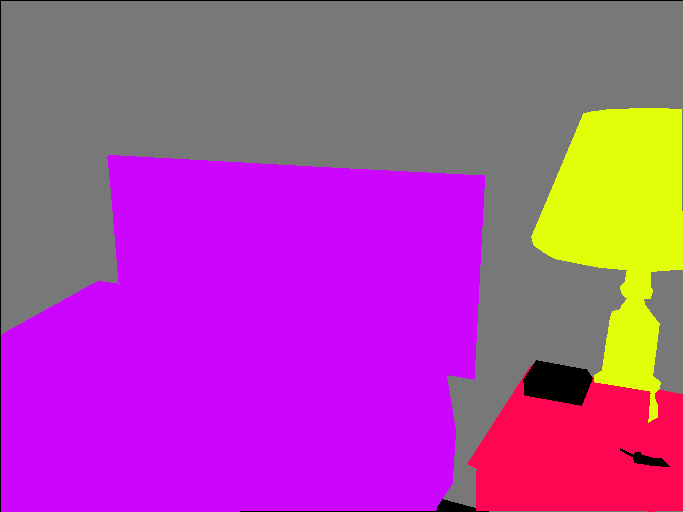}\\
		\includegraphics[width=\linewidth]{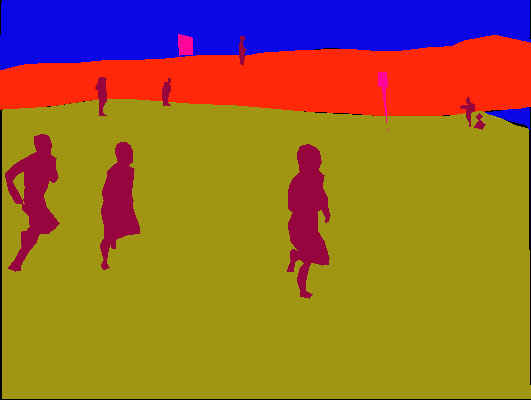}\\
		\caption{GT}
    \end{subfigure}
    \hfill
	\begin{subfigure}[b]{0.24\linewidth}
		\includegraphics[width=\linewidth]{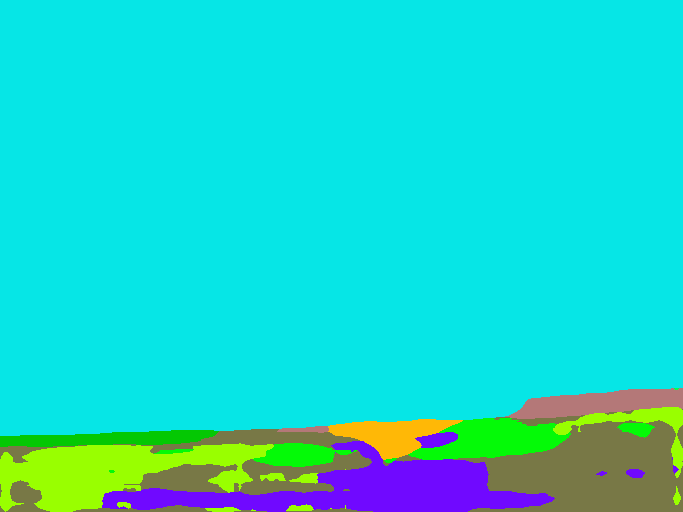}\\
		\includegraphics[width=\linewidth]{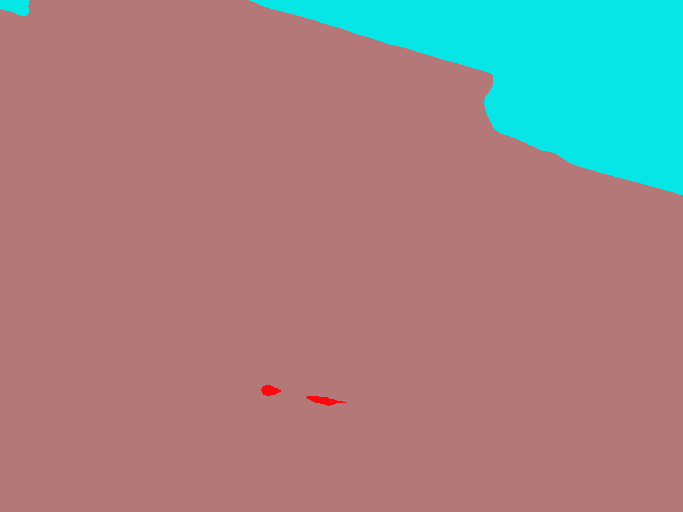}\\
		\includegraphics[width=\linewidth]{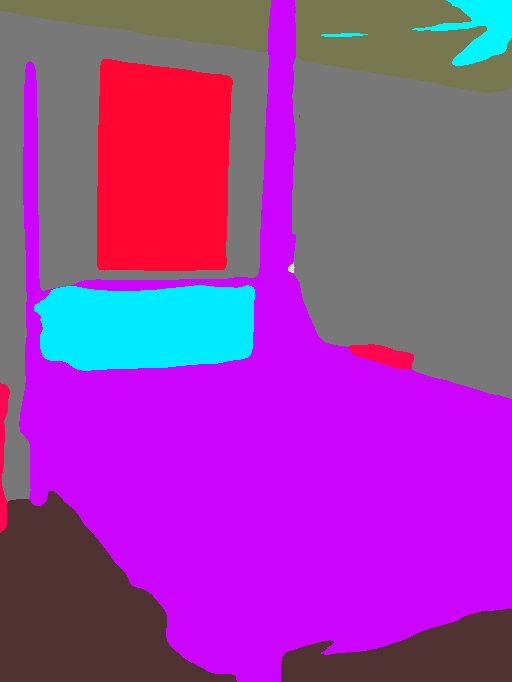}\\
		\includegraphics[width=\linewidth]{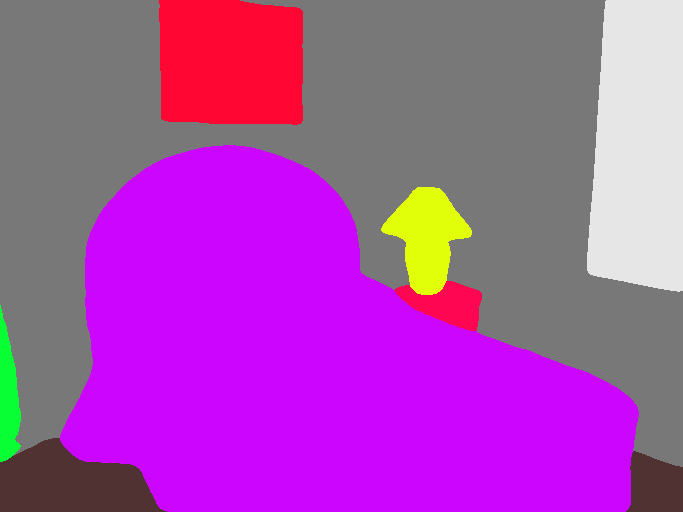}\\
		\includegraphics[width=\linewidth]{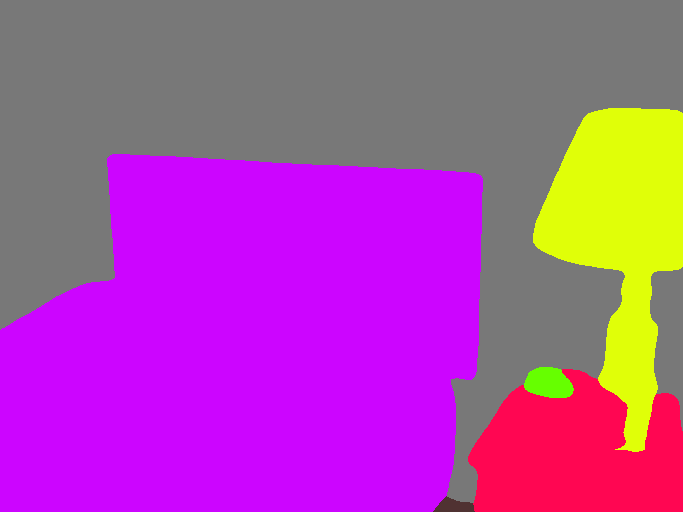}\\
		\includegraphics[width=\linewidth]{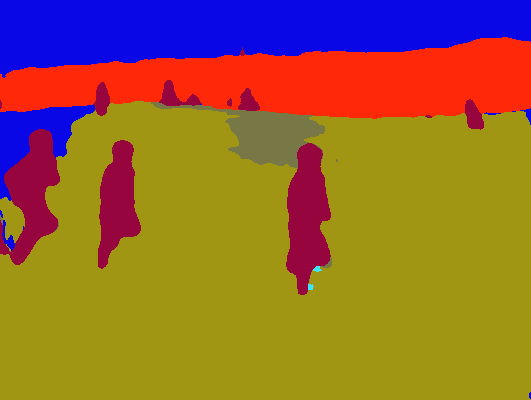}\\
		\caption{EncNet}
    \end{subfigure}
    \hfill
    \begin{subfigure}[b]{0.24\linewidth}
		\includegraphics[width=\linewidth]{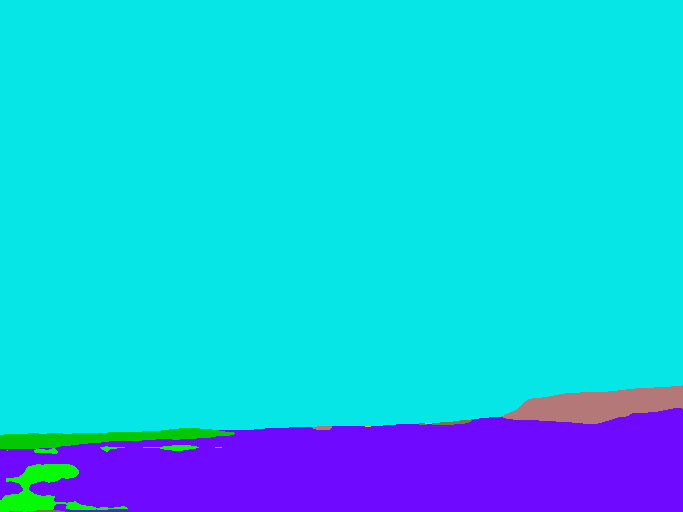}\\
		\includegraphics[width=\linewidth]{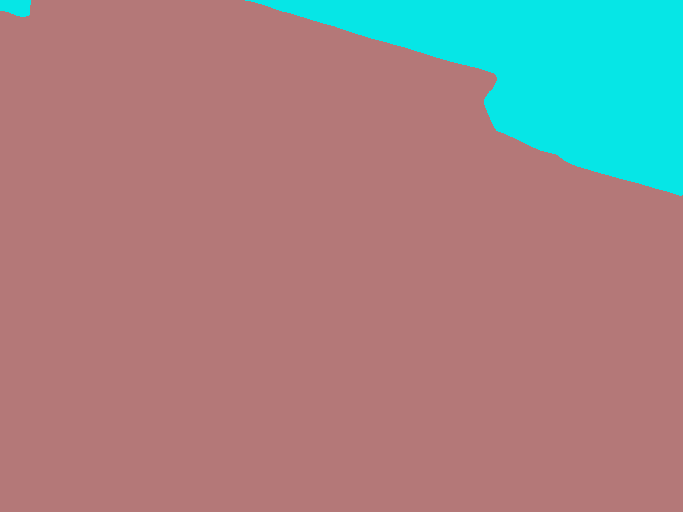}\\
		\includegraphics[width=\linewidth]{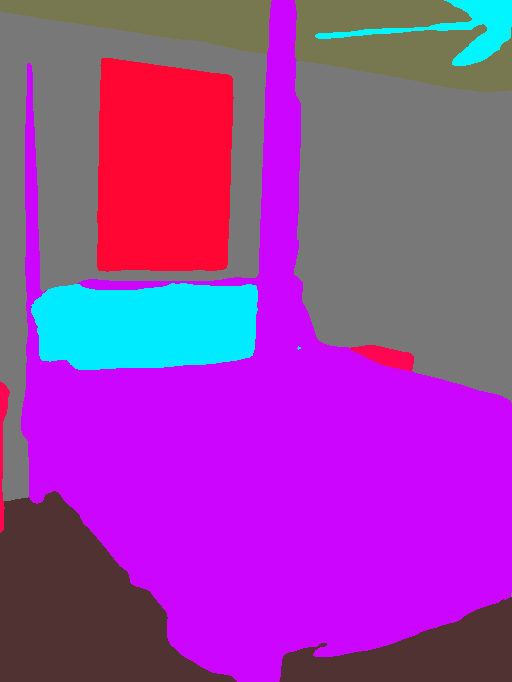}\\
		\includegraphics[width=\linewidth]{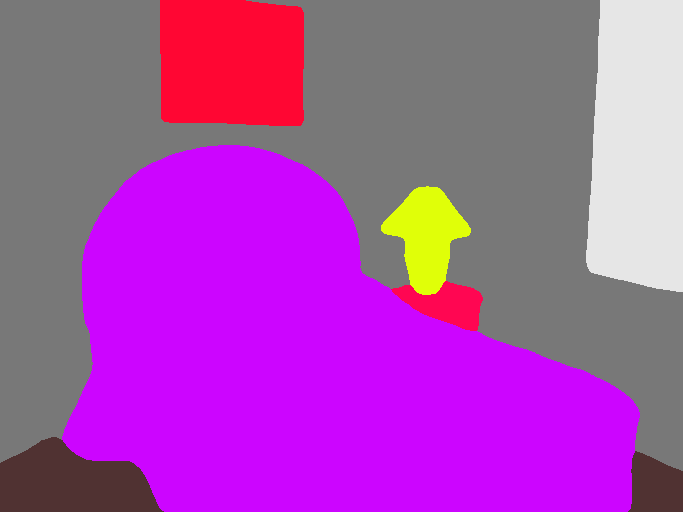}\\
		\includegraphics[width=\linewidth]{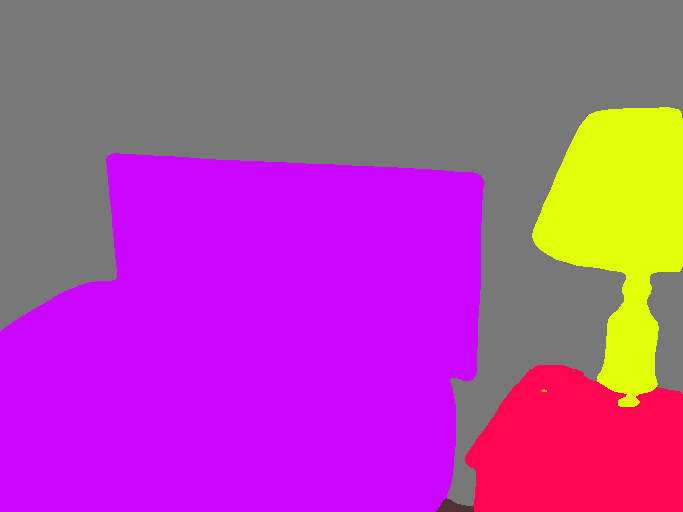}\\
		\includegraphics[width=\linewidth]{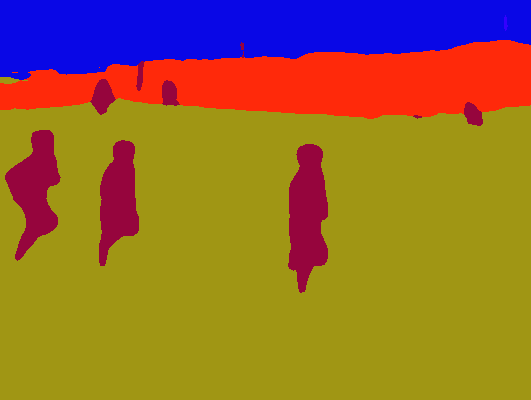}\\
		\caption{Ours}
    \end{subfigure}
\end{center}
	\caption{Visual results of our method (ResNet-101) on the ADE20K \textit{val} set. Best viewed in color.}
\end{figure*}

\end{document}